\pdfoutput=1
\documentclass[11pt,a4paper]{article}
\usepackage[hyperref]{emnlp2020}
\usepackage{times}
\usepackage{latexsym}

\usepackage{microtype}

\usepackage[absolute,overlay]{textpos}

\usepackage{arydshln}
\usepackage{graphicx}
\usepackage{booktabs}
\usepackage{amssymb}  
\usepackage{subcaption}
\usepackage{amsmath} 

\newcommand{\model}[1]{\textsc{MAD-X}}
\newcommand{\xlmr}[1]{\textsc{XLM-R}}
\newcommand{\mlmsrc}[1]{\textsc{XLM-R$^{Base}$ MLM-src}}
\newcommand{\mlmtrg}[1]{\textsc{XLM-R$^{Base}$ MLM-trg}}

\newcommand{\wann}[1]{WikiANN}

\aclfinalcopy 
\def\aclpaperid{1803} 

\title{MAD-X: An Adapter-Based Framework for\\Multi-Task Cross-Lingual Transfer}

\author{Jonas Pfeiffer$^{1}$, Ivan Vuli\'{c}$^{2}$, {\bf Iryna Gurevych$^{1}$, Sebastian Ruder$^{3}$ } \\
$^1$Ubiquitous Knowledge Processing Lab, 
  Technical University of Darmstadt \\
$^2$Language Technology Lab, University of Cambridge \hspace{0.5em} \\
$^3$DeepMind \\
\texttt{pfeiffer@ukp.tu-darmstadt.de} \\
}

\date{}

\begin{document}
\maketitle
\begin{abstract}
The main goal behind state-of-the-art pretrained multilingual models such as multilingual BERT and XLM-R is enabling and bootstrapping NLP applications in \textit{low-resource languages} through zero-shot or few-shot cross-lingual transfer. However, due to limited model capacity, their transfer performance is the weakest exactly on such low-resource languages and languages \textit{unseen} during pretraining. We propose \textbf{\model{}}, an adapter-based framework that enables high portability and parameter-efficient transfer to arbitrary tasks and languages by learning modular language and task representations. In addition, we introduce a novel invertible adapter architecture and a strong baseline method for adapting a pretrained multilingual model to a new language. \model{} outperforms the state of the art in cross-lingual transfer across a representative set of typologically diverse languages on named entity recognition and causal commonsense reasoning, and achieves competitive results on question answering. Our code and adapters are available at \href{https://AdapterHub.ml}{AdapterHub.ml}.
\end{abstract}

\begin{textblock*}{4cm}(1.15cm,-0.3cm) 
\includegraphics[width=5cm]{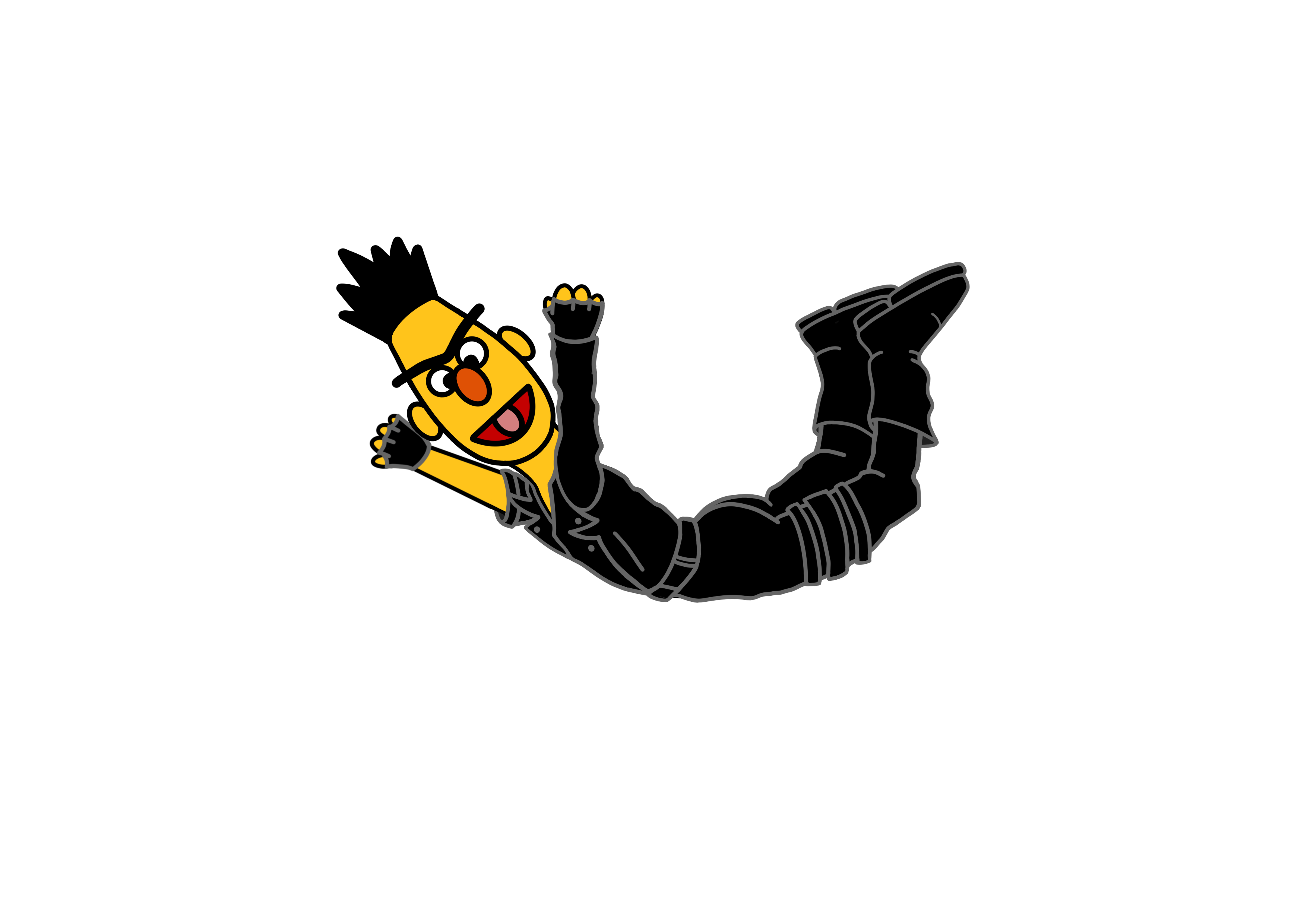}
\end{textblock*}

\section{Introduction}
\label{s:introduction}
Current deep pretrained multilingual models \cite{Devlin2019bert,Lample2019xlm} achieve state-of-the-art results on cross-lingual transfer, but do not have enough capacity to represent all languages. Evidence for this is the importance of the vocabulary size \cite{Artetxe2020cross-lingual} and \textit{the curse of multilinguality} \cite{Conneau2020xlm-r}, a trade-off between language coverage and model capacity. Scaling up a model to cover all of the world's 7,000+ languages is prohibitive. At the same time, limited capacity is an issue even for high-resource languages where state-of-the-art multilingual models underperform their monolingual variants \cite{Eisenschlos2019multifit,Virtanen:2019arxiv,Nozza:2020arxiv}, and performance decreases further with lower-resource languages covered by the pretrained models. Moreover, the model capacity issue is arguably most severe for languages that were not included in the training data at all, and pretrained models perform poorly on those languages \cite{Ponti2020factorization}.

In this paper, we propose \textsc{M}ultiple \textsc{AD}apters for Cross-lingual transfer (\textbf{\model{}}), a modular framework that leverages a small number of extra parameters to address the fundamental capacity issue that limits pretrained multilingual models. Using a state-of-the-art multilingual model as the foundation, we adapt the model to arbitrary tasks and languages by learning modular language- and task-specific representations \textit{via adapters} \cite{Rebuffi2017adapters,Houlsby2019adapters}, small bottleneck layers inserted between a model's weights.

In particular, using a recent efficient adapter variant \cite{Pfeiffer2020adapterfusion, rueckle2020adapterdrop}, we train \textbf{1)} \textit{language-specific adapter modules} via masked language modelling (MLM) on unlabelled target language data, and \textbf{2)} \textit{task-specific adapter modules} via optimising a target task on labelled data in any source language. Task and language adapters are stacked as in Figure~\ref{fig:Mad-X_full}, allowing us to adapt the pretrained multilingual model also to languages that are not covered in the model's (pre)training data by substituting the target language adapter at inference.

In order to deal with a mismatch between the shared multilingual vocabulary and target language vocabulary, we propose \textit{invertible adapters}, a new type of adapter that is well suited to performing MLM in another language. Our framework goes beyond prior work on using adapters for cross-lingual transfer \cite{Bapna2019adapters,Artetxe2020cross-lingual} by enabling adaptation to languages \textit{unseen} during pretraining and without learning expensive language-specific token-level embeddings.

We compare \model{} against state-of-the-art cross-lingual transfer methods on the standard \wann{} NER dataset \cite{Pan2017wikiann,Rahimi2019massively} and the XCOPA dataset \cite{Ponti2020xcopa} for causal commonsense reasoning, relying on a representative set of typologically diverse languages which includes high-resource, low-resource, as well as languages unseen by the pretrained model. \model{} outperforms the baselines on seen and unseen high-resource and low-resource languages. On the high-resource languages of the challenging XQuAD QA dataset \cite{Artetxe2020cross-lingual}, our framework achieves competitive performance while being more parameter-efficient.

Another contribution of our work is a simple method of adapting a pretrained multilingual model to a new language, which outperforms the standard setting of transferring a model only from labelled source language data. %

In sum, our contributions are as follows. \textbf{1)}~We propose \model{}, a modular framework that mitigates the curse of multilinguality and adapts a multilingual model to arbitrary tasks and languages. Both code and adapter weights are integrated into the \href{https://AdapterHub.ml}{AdapterHub.ml} repository \cite{pfeiffer2020AdapterHub}.\footnote{\href{https://github.com/Adapter-Hub/adapter-transformers}{https://github.com/Adapter-Hub/adapter-transformers}} \textbf{2)} We propose invertible adapters, a new adapter variant for cross-lingual MLM. \textbf{3)} We demonstrate strong performance and robustness of \model{} across diverse languages and tasks. \textbf{4)} We propose a simple and more effective baseline method for adapting a pretrained multilingual model to target languages. \textbf{5)} We shed light on the behaviour of current methods on 
languages that are unseen during multilingual pretraining.

\section{Related Work}

\noindent \textbf{Cross-lingual Representations}\hspace{0.3mm} Research in modern cross-lingual NLP is increasingly focused on learning general-purpose cross-lingual representations that can be applied to many tasks, first on the word level \cite{mikolov2013exploiting,gouws2015bilbowa,Glavas:2019acl,Ruder2019survey,Wang:2020iclr} and later on the full-sentence level \cite{Devlin2019bert,Lample2019xlm,Cao:2020iclr}. More recent models such as multilingual BERT \cite{Devlin2019bert}---large Transformer \cite{Vaswani2017transformer} models pretrained on large amounts of multilingual data---have been observed to perform surprisingly well when transferring to other languages \cite{Pires2019,Wu2019beto,Wu2020emerging} and the current state-of-the-art model, XLM-R is competitive with the performance of monolingual models on the GLUE benchmark \cite{Conneau2020xlm-r}. Recent studies \cite{Hu2020xtreme}, however, indicate that state-of-the-art models such as XLM-R still perform poorly on cross-lingual transfer across many language pairs. The main reason behind such poor performance is the current lack of capacity in the model to represent all languages equally in the vocabulary and representation space \cite{Bapna2019adapters,Artetxe2020cross-lingual,Conneau2020xlm-r}.

\vspace{1.8mm}
\noindent \textbf{Adapters}\hspace{0.3mm} Adapter modules have been originally studied in computer vision tasks where they have been restricted to convolutions and used to adapt a model for multiple domains \cite{Rebuffi2017adapters,Rebuffi2018}. In NLP, adapters have been mainly used for parameter-efficient and quick fine-tuning of a base pretrained Transformer model to new tasks  \cite{Houlsby2019adapters,Cooper2019adapters} and new domains \cite{Bapna2019adapters}, avoiding catastrophic forgetting \cite{McCloskey:1989,Santoro:2016arxiv}. \citet{Bapna2019adapters} also use adapters to fine-tune and recover performance of a multilingual NMT model on high-resource languages, but their approach cannot be applied to languages that were not seen during pretraining. \citet{Artetxe2020cross-lingual} employ adapters to transfer a pretrained monolingual model to an unseen language but rely on learning new token-level embeddings, which do not scale to a large number of languages. \citet{Pfeiffer2020adapterfusion} combine the information stored in multiple adapters for more robust transfer learning between monolingual tasks. In their contemporaneous work, \citet{ustun2020udapter} generate adapter parameters from language embeddings for multilingual dependency parsing.
 
\section{Multilingual Model Adaptation for Cross-lingual Transfer}
\label{s:adaptation}

\noindent \textbf{Standard Transfer Setup}\hspace{0.3mm}
The standard way of performing cross-lingual transfer with a state-of-the-art large multilingual model such as multilingual BERT or XLM-R is 1) to fine-tune it on labelled data of a downstream task in a source language and then 2) apply it directly to perform inference in a target language \cite{Hu2020xtreme}. A downside of this setting is that the multilingual initialisation balances \emph{many} languages. It is thus not suited to excel at a specific language at inference time. We propose a simple method to ameliorate this issue by allowing the model to additionally adapt to the particular target language.
 
\vspace{1.8mm}
\noindent \textbf{Target Language Adaptation} \hspace{0.3mm}
Similar to fine-tuning monolingual models on the task domain \cite{Howard2018ulmfit}, we propose to fine-tune a pretrained multilingual model via MLM on unlabelled data of the target language prior to task-specific fine-tuning in the source language. A disadvantage of this approach is that it no longer allows us to evaluate the same model on multiple target languages as it biases the model to a specific target language. However, this approach might be preferable if we only care about performance in a specific (i.e., fixed) target language. We find that target language adaptation results in improved cross-lingual transfer performance over the standard setting (\textsection \ref{sec:results}). In other words, it does not result in catastrophic forgetting of the multilingual knowledge already available in the pretrained model that enables the model to transfer to other languages. In fact, experimenting with methods that explicitly try to prevent catastrophic forgetting \cite{Wiese2017} led to worse performance in our experiments.

Nevertheless, the proposed simple adaptation method inherits the fundamental limitation of the pretrained multilingual model and the standard transfer setup: the model's limited capacity hinders effective adaptation to low-resource and unseen languages. In addition, fine-tuning the full model does not scale well to many tasks or languages.

 \begin{figure}[!t] 
\centering
\includegraphics[width=1.0\linewidth]{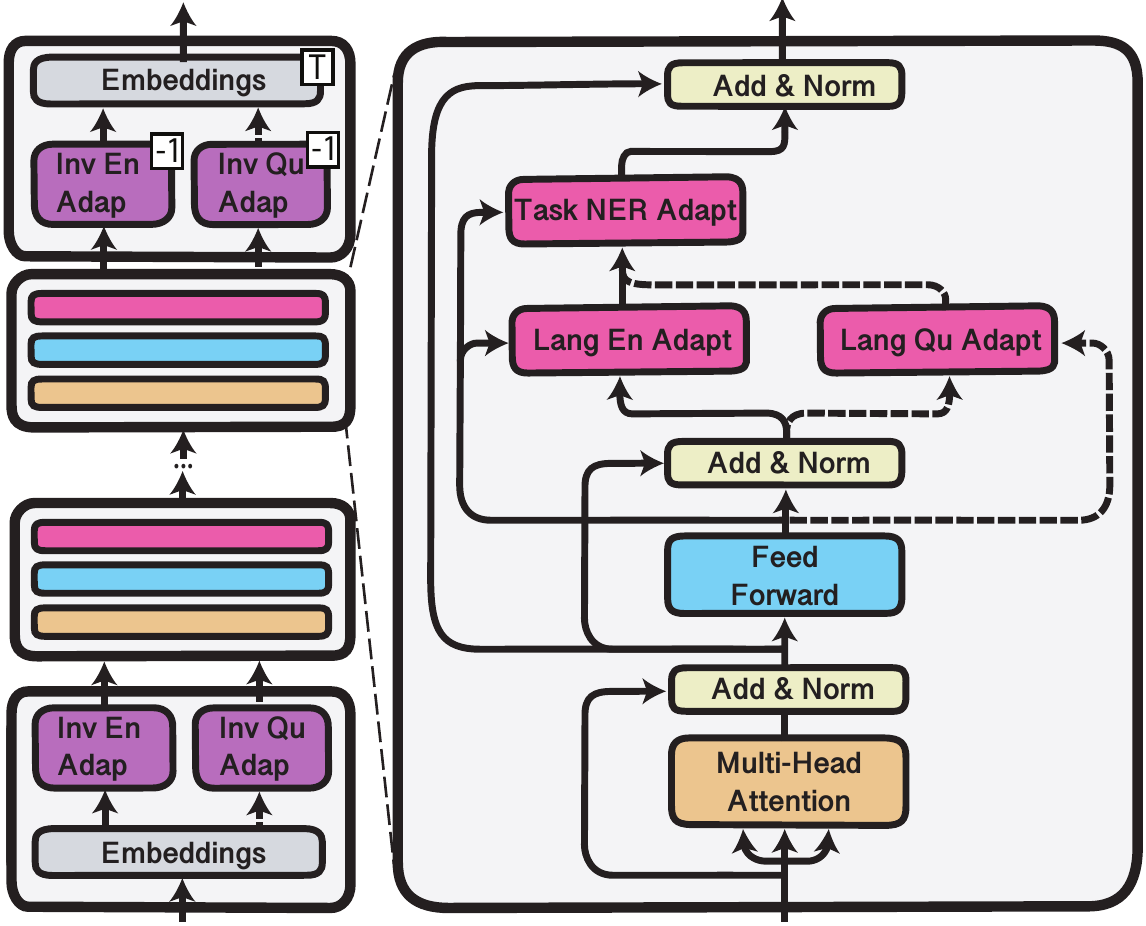}
\caption{The \model{} framework inside a Transformer model. Input embeddings are fed into the invertible adapter whose inverse is fed into the tied output embeddings. Language and task adapters are added to each Transformer layer. Language adapters and invertible adapters are trained via masked language modelling (MLM) while the pretrained multilingual model is kept frozen. Task-specific adapters are stacked on top of source language adapters when training on a downstream task such as NER (full lines). During zero-shot cross-lingual transfer, source language adapters are replaced with target language adapters (dashed lines).}
\label{fig:Mad-X_full}
\end{figure}

\section{Adapters for Cross-lingual Transfer}
 
Our \model{} framework addresses these deficiencies and can be used to effectively adapt an existing pretrained multilingual model to other languages. The framework comprises three types of adapters: language, task, and invertible adapters. As in previous work \cite{Rebuffi2017adapters,Houlsby2019adapters}, adapters are trained while keeping the parameters of the pretrained multilingual model fixed. Our framework thus enables learning language and task-specific transformations in a modular and parameter-efficient way. We show the full framework as part of a standard Transformer model in Figure~\ref{fig:Mad-X_full} and describe the three adapter types.  
\subsection{Language Adapters}
\label{sec:transformer_adapters}

For learning language-specific transformations, we employ a recent efficient adapter architecture proposed by \citet{Pfeiffer2020adapterfusion}. Following \citet{Houlsby2019adapters} they define the interior of the adapter to be a simple down- and up-projection combined with a residual connection.\footnote{\citet{Pfeiffer2020adapterfusion} perform an extensive hyperparameter search over adapter positions, activation functions, and residual connections within each Transformer layer. They arrive at an architecture variant that performs on par with that of \citet{Houlsby2019adapters}, while being more efficient.}  The language adapter $\mathsf{LA}_l$ at layer $l$ consists of a down-projection $\textbf{D} \in \mathbb{R}^{h \times d}$ where $h$ is the hidden size of the Transformer model and $d$ is the dimension of the adapter, followed by a $\mathsf{ReLU}$ activation and an up-projection $\textbf{U}\in \mathbb{R}^{d \times h}$ at every layer $l$:
\begin{equation}
\mathsf{LA}_l(\textbf{h}_l, \textbf{r}_l) = \textbf{U}_l(\mathsf{ReLU}(\textbf{D}_l(\textbf{h}_l))) + \textbf{r}_l
\end{equation}
where $\textbf{h}_l$ and $\textbf{r}_l$ are the Transformer hidden state and the residual at layer $l$, respectively. The residual connection $\textbf{r}_l$ is the output of the Transformer's feed-forward layer whereas $\textbf{h}_l$ is the output of the subsequent layer normalisation (see Figure~\ref{fig:Mad-X_full}).

We train language adapters on unlabelled data of a language using MLM, which encourages them to learn transformations that make the pretrained multilingual model more suitable for a specific language. During task-specific training with labelled data, we use the language adapter of the corresponding source language, which is kept fixed. In order to perform zero-shot transfer to another language, we simply replace the source language adapter with its target language component. For instance, as illustrated in Figure~\ref{fig:Mad-X_full}, we can simply replace a language-specific adapter trained for English with a language-specific adapter trained for Quechua at inference time. This, however, requires that the underlying multilingual model does not change during fine-tuning on the downstream task. In order to ensure this, we additionally introduce task adapters that capture task-specific knowledge.

\begin{figure}[!t]
    \centering
    \begin{subfigure}[t]{0.466\linewidth}
        \centering
        \includegraphics[width=0.98\linewidth]{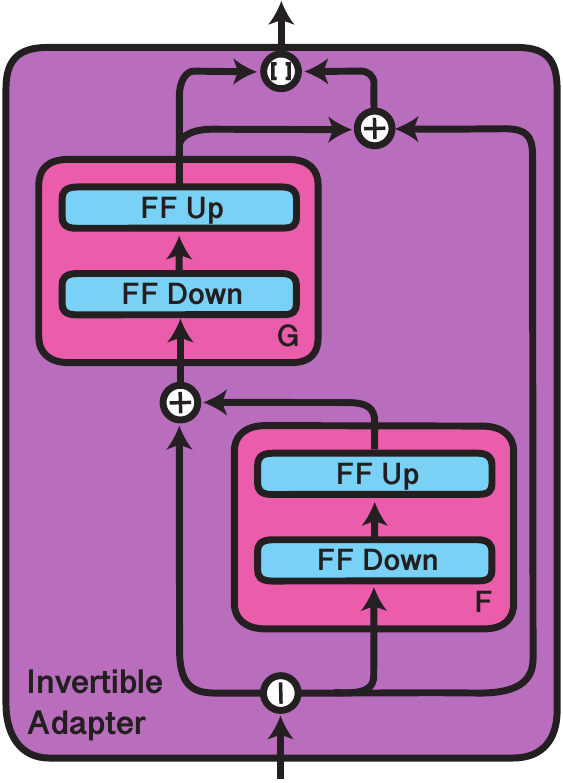}
        \caption{The invertible adapter}
        \label{fig:invertible-adapter}
    \end{subfigure}
    \hspace{2mm}
    \begin{subfigure}[t]{0.475\linewidth}
        \centering
        \includegraphics[width=0.98\linewidth]{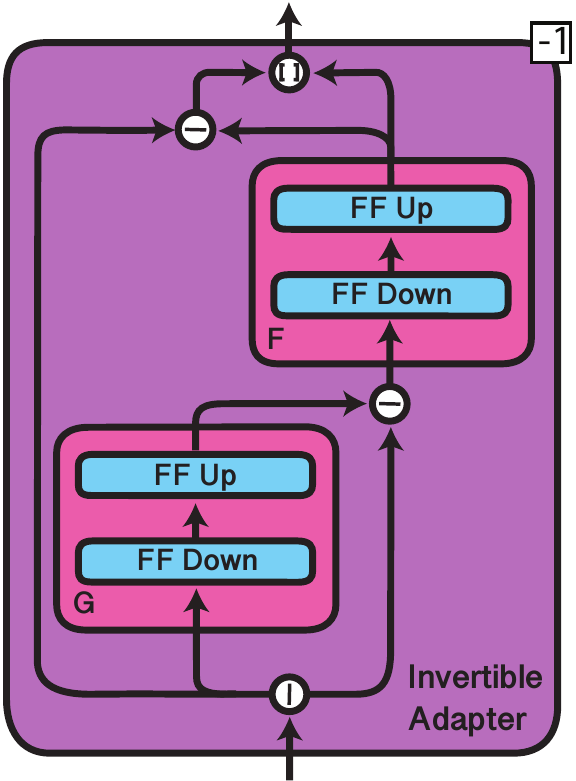}
        \caption{The inversed adapter}
        \label{fig:inverse-invertible-adapter}
    \end{subfigure}
    \vspace{-1.5mm}
    \caption{The invertible adapter (a) and its inverse (b). The input is split and transformed by projections $F$ and $G$, which are coupled in an alternating fashion. $|$ indicates the splitting of the input vector, and $[\,\,]$ indicates the concatenation of two vectors. $+$ and $-$ indicate element-wise addition and subtraction, respectively.  }
    \vspace{-1.5mm}
\label{fig:invertible-adapter-variants}
\end{figure}

\subsection{Task Adapters}

Task adapters $\mathsf{TA}_l$ at layer $l$ have the same architecture as language adapters. They similarly consist of a down-projection $\textbf{D} \in \mathbb{R}^{h \times d}$, a $\mathsf{ReLU}$ activation, followed by an up-projection. They are stacked on top of the language adapters and thus receive the output of the language adapter $\mathsf{LA}_l$ as input, together with the residual $\mathbf{r}_l$ of the Transformer's feed-forward layer\footnote{Initial experiments showed that this residual connection performs better than one to the output of the language adapter.}:
\begin{equation}
\mathsf{TA}_l(\textbf{h}_l, \textbf{r}_l) = \textbf{U}_l(\mathsf{ReLU}(\textbf{D}_l(\mathsf{LA}_l))) + \textbf{r}_l
\end{equation}
\noindent The output of the task adapter is then passed to another layer normalisation component. Task adapters are the only parameters that are updated when training on a downstream task (e.g., NER) and aim to capture knowledge that is task-specific but generalises across languages.

\subsection{Invertible Adapters}
\label{sec:invertible_adapters}

The majority of the ``parameter budget'' of pretrained multilingual models is spent on token embeddings of the shared multilingual vocabulary. Despite this, they underperform on low-resource languages \cite{Artetxe2020cross-lingual,Conneau2020xlm-r}, and are bound to fare even worse for languages not covered by the model's training data.  

In order to mitigate this mismatch between multilingual and target language vocabulary, we propose invertible adapters. They are stacked on top of the embedding layer while their respective inverses precede the output embedding layer (see Figure~\ref{fig:Mad-X_full}).  As input and output embeddings are tied in multilingual pretrained models, invertibility allows us to leverage the same set of parameters for adapting both input and output representations. This is crucial as the output embeddings, which get discarded during task-specific fine-tuning might otherwise overfit to the pretraining task. 

To ensure this invertibility, we employ Non-linear Independent Component Estimation \cite[NICE;][]{Dinh2014NICE}. NICE enables the invertibility of arbitrary non-linear functions through a set of coupling operations \cite{Dinh2014NICE}.
For the invertible adapter, we split the input embedding vector $\textbf{e}_i$ of the $i$-th token into two vectors of equal dimensionality $\textbf{e}_{1,i}, \textbf{e}_{2,i} \in \mathbb{R}^{h/2}$.\footnote{For brevity, we further leave out the dependency on $i$.} For two arbitrary non-linear function $F$ and $G$, the forward pass through our invertible adapter $A_{inv}()$ is:
\begin{equation}
\begin{split}
    &\textbf{o}_1  =F(\textbf{e}_2) + \textbf{e}_1;\hspace{2mm} \textbf{o}_2 =  G(\textbf{o}_1) + \textbf{e}_2 \\
    &\textbf{o} = [\textbf{o}_1 , \textbf{o}_2]
\end{split}
\end{equation}
where $\textbf{o}$ is the output of the invertible adapter $A_{inv}$ and $[\cdot,\cdot]$ indicates concatenation of two vectors. 

Correspondingly, the inverted pass through the adapter, thus $A_{inv}^{-1}$, is computed as follows:
\begin{equation}
\begin{split}
    & \textbf{e}_2 = \textbf{o}_2 - G(\textbf{o}_1);\hspace{2mm} \textbf{e}_1 = \textbf{o}_1 - F(\textbf{e}_2) \\
    & \textbf{e} = [\textbf{e}_1 , \textbf{e}_2].
\end{split}
\end{equation}
$\textbf{e}$ is the output of $A_{Inv}^{-1}()$. For the non-linear transformations $F$ and $G$, we use similar down- and up-projections as for the language and task adapters:
\vspace{-1mm}
\begin{equation}
\begin{split}
F(\textbf{x}) &  = \textbf{U}_F(\mathsf{ReLU}(\textbf{D}_F(\textbf{x})))  \\
G(\textbf{x}) & = \textbf{U}_G(\mathsf{ReLU}(\textbf{D}_G(\textbf{x}))).
\end{split}
\end{equation}
where $\textbf{D}_F, \textbf{D}_G \in \mathbb{R}^{\frac{h}{4} \times \frac{h}{2}}$ and  $\textbf{U}_F, \textbf{U}_G \in \mathbb{R}^{\frac{h}{2} \times \frac{h}{4}}$ and $\textbf{x}$ is a placeholder for $\textbf{e}_1, \textbf{e}_2, \textbf{o}_1$ and $\textbf{o}_2$. We illustrate the complete architecture of the invertible adapter and its inverse in Figure~\ref{fig:invertible-adapter-variants}. 

The invertible adapter has a similar function to the language adapter, but aims to capture token-level language-specific transformations. As such, it is trained together with the language adapters using MLM on unlabelled data of a specific language. During task-specific training we use the fixed invertible adapter of the source language, and replace it with the target-language invertible during zero-shot transfer. Importantly, our invertible adapters are much more parameter-efficient compared to the approach of \citet{Artetxe2020cross-lingual}, which learns separate token embeddings for every new language.

\vspace{1.8mm}
\noindent \textbf{An Illustrative Example}\hspace{1.3mm}
We  provide a brief walk-through example from Figure~\ref{fig:Mad-X_full}. Assuming English (\textit{En}) as the source language and Quechua (\textit{Qu}) as the target language, we first pretrain invertible adapters $A_{Inv}^{En}$ and $A_{Inv}^{Qu}$, and language adapters $A_{Lang}^{En}$ and $A_{Lang}^{Qu}$ with MLM for which the output of the last  layer is passed through $A_{Inv}^{En}{}^{-1}$. We then train a task adapter for the NER task $A_{Task}^{NER}$ on the English NER training set. During training, embeddings are passed through $A_{Inv}^{En}$. At every layer of the model the data is first passed through the fixed $A_{Lang}^{En}$ and then into the NER adapter $A_{Task}^{NER}$.  
For zero-shot inference, the English invertible and language adapters $A_{Inv}^{En}$ and $A_{Lang}^{En}$ are  replaced with their Quechua counterparts $A_{Inv}^{Qu}$ and $A_{Lang}^{Qu}$ while the data is still passed through the NER task adapter $A_{Task}^{NER}$.

\section{Experiments}

\begin{table}[]
\centering
\def\arraystretch{1.05}
\resizebox{\columnwidth}{!}{%
\begin{tabular}{l l l r c}
\toprule
Language & \begin{tabular}[c]{@{}l@{}}ISO\\ code\end{tabular} & \begin{tabular}[c]{@{}l@{}}Language\\ family\end{tabular} & \begin{tabular}[r]{@{}r@{}}\# of Wiki \\ articles\end{tabular} & \begin{tabular}[c]{@{}l@{}}Covered\\ by SOTA?\end{tabular} \\ \midrule
English & en & Indo-European & 6.0M & \checkmark \\
Japanese & ja & Japonic & 1.2M & \checkmark \\
Chinese & zh & Sino-Tibetan & 1.1M & \checkmark \\
Arabic & ar & Afro-Asiatic & 1.0M & \checkmark \\
\hdashline
Javanese & jv & Austronesian & 57k & \checkmark \\
Swahili & sw & Niger-Congo & 56k & \checkmark \\
Icelandic & is & Indo-European & 49k & \checkmark \\
Burmese & my & Sino-Tibetan & 45k & \checkmark \\
\hdashline
Quechua & qu & Quechua & 22k &  \\
Min Dong & cdo & Sino-Tibetan & 15k &  \\
Ilokano & ilo & Austronesian & 14k &  \\
Mingrelian & xmf & Kartvelian & 13k &  \\
\hdashline
Meadow Mari & mhr & Uralic & 10k &  \\
Maori & mi & Austronesian & 7k &  \\
Turkmen & tk & Turkic & 6k &  \\
Guarani & gn & Tupian & 4k & \\ \bottomrule
\end{tabular}%
}
\caption{Languages in our NER evaluation.}
\label{tab:ner-languages}
\end{table}

\noindent \textbf{Data}\hspace{0.3mm} We conduct experiments on three tasks: Named entity recognition (NER), question answering (QA), and causal commonsense reasoning (CCR). For NER, we use the WikiANN \cite{Pan2017wikiann} dataset, which was partitioned into train, development, and test portions by \citet{Rahimi2019massively}. For QA, we employ the XQuAD dataset \cite{Artetxe2020cross-lingual}, a cross-lingual version of SQuAD \cite{Rajpurkar2016squad}. For CCR, we rely on XCOPA \cite{Ponti2020xcopa}, a cross-lingual version of COPA \cite{roemmele2011choice}.

\vspace{1.8mm}
\noindent \textbf{Languages}\hspace{0.3mm} 
The partitioned version of WikiANN covers 176 languages. In order to obtain a comprehensive comparison to state-of-the-art cross-lingual methods under different evaluation conditions, we select languages based on: \textbf{a)} variance in data availability (by selecting languages with a range of respective Wikipedia sizes); \textbf{b)} their presence in pretrained multilingual models; more precisely, whether data in the particular language was included in the pretraining data of both multilingual BERT and XLM-R or not; and \textbf{c)} typological diversity to ensure that different language types and families are covered. In total, we can discern four categories in our language set: \textbf{1)} high-resource languages and \textbf{2)} low-resource languages covered by the pretrained SOTA multilingual models (i.e., by mBERT and XLM-R); as well as \textbf{3)} low-resource languages and \textbf{4)} truly low-resource languages not covered by the multilingual models. We select four languages from different language families for each category. We highlight characteristics of the 16 languages from 11 language families in Table~\ref{tab:ner-languages}.

We evaluate on all possible language pairs (i.e., on the Cartesian product), using each language as a source language with every other language (including itself) as a target language. This subsumes both the standard \textit{zero-shot cross-lingual transfer setting} \cite{Hu2020xtreme} as well as the standard \textit{monolingual in-language} setting.

For CCR and QA, we evaluate on the 12 and 11 languages provided in XCOPA and XQuAD respectively, with English as source language. XCOPA contains a typologically diverse selection of languages including two languages (Haitian Creole and Quechua) that are unseen by our main model. XQuAD comprises slightly less typologically diverse languages that are mainly high-resource.

\begin{table*}[]
\centering
\def\arraystretch{1.0}
\resizebox{\textwidth}{!}{%
\begin{tabular}{lllllllll:llllllll | l}
\toprule
Model & en & ja & zh & ar & jv & sw & is & my & qu & cdo & ilo & xmf & mi & mhr & tk & gn & avg \\
\midrule
\xlmr{}$^{Base}$	&   44.2	&	38.2	&	40.4	&	36.4	&	37.4	&	42.8	&	47.1	&	\textbf{26.3}	&	27.4	&	18.1	&	28.8	&	\textbf{35.0}	&	16.7	&	\textbf{31.7}	&	20.6	&	\textbf{31.2}	&	32.6 \\
\mlmsrc{}	& 39.5	&	45.2	&	34.7	&	17.7	&	34.5	&	35.3	&	43.1	&	20.8	&	26.6	&	21.4	&	28.7	&	22.4	&	18.1	&	25.0	&	27.6	&	24.0	&	29.0 \\
\mlmtrg{} & 54.8	&	\textbf{47.4}	&	\textbf{54.7}	&	51.1	&	38.7	&	48.1	&	53.0	&	20.0	&	29.3	&	16.6	&	27.4	&	24.7	&	15.9	&	26.4	&	26.5	&	28.5	&	35.2\\
\midrule
\model{}$^{Base}$ $\:$ -- \textsc{lad} -- \textsc{inv}	& 44.5	&	38.6	&	40.6	&	42.8	&	32.4	&	43.1	&	48.6	&	23.9	&	22.0	&	10.6	&	23.9	&	27.9	&	13.2	&	24.6	&	18.8	&	21.9	&	29.8 \\
\model{}$^{Base}$ $\:$ -- \textsc{inv}	&   52.3	&	46.0	&	46.2	&	56.3	&	\textbf{41.6}	&	48.6	&	52.4	&	23.2	&	32.4	&	\textbf{27.2}	&	30.8	&	33.0	&	\textbf{23.5}	&	29.3	&	30.4	&	28.4	&	37.6 \\ 
\model{}$^{Base}$ $\:$  	&	\textbf{55.0}	&	46.7	&	47.3	&	\textbf{58.2}	&	39.2	&	\textbf{50.4}	&	\textbf{54.5}	&	24.9	&	\textbf{32.6}	&	24.2	&	\textbf{33.8}	&	34.3	&	16.8	&	\textbf{31.7}	&	\textbf{31.9}	&	30.4	&	\textbf{38.2} \\ 
\midrule
\midrule

mBERT	&	48.6	&	50.5	&	50.6	&	50.9	&	45.3	&	48.7	&	51.2	&	17.7	&	31.8	&	20.7	&	33.3	&	26.1	&	\textbf{20.9}	&	31.3	&	\textbf{34.8}	&	\textbf{30.9}	&	37.1 \\
\model{}$^{mBERT}$	&	\textbf{52.8}	&	\textbf{53.1}	&	\textbf{53.2}	&	\textbf{55.5}	&	\textbf{46.3}	&	\textbf{50.9}	&	\textbf{51.4}	&	\textbf{21.0}	&	\textbf{37.7}	&	\textbf{22.1}	&	\textbf{35.0}	&	\textbf{30.0}	&	18.6	&	\textbf{31.8}	&	33.0	&	25.1	&	\textbf{38.6} \\

\midrule
\midrule

XLM-R$^{Large}$	&	47.10	&	46.52	&	46.43	&	45.15	&	39.21	&	43.96	&	48.69	&	\textbf{26.18}	&	26.39	&	15.12	&	22.80	&	\textbf{33.67}	&	19.86	&	27.70	&	29.56	&	\textbf{33.78}	&	34.6 \\
\model{}$^{Large}$	&	\textbf{56.30} &	\textbf{53.37} &	\textbf{55.6}	& \textbf{59.41}	&	\textbf{40.40}	&	\textbf{50.57}	&	\textbf{53.22}	&	24.55	&	\textbf{33.89}	&	\textbf{26.54}	&	\textbf{30.98}	&	33.37	&	\textbf{24.31}	&	\textbf{28.03}	&	\textbf{30.82}	&	26.38 & \textbf{39.2} \\

\bottomrule
\end{tabular}%
}
\caption{
NER F1 scores averaged over all 16 target languages when transferring from each source language (i.e. the columns are source languages). The vertical dashed line distinguishes between languages seen in multilingual pretraining and the unseen ones (see also Table~\ref{tab:ner-languages}).}
\label{tab:main-results}
\end{table*}

\subsection{Baselines}
The baseline models are based on different approaches to multilingual model adaptation for cross-lingual transfer, discussed previously in \S\ref{s:adaptation}.

\vspace{1.8mm}
\noindent \textbf{\xlmr{}}\hspace{0.3mm}
The main model we compare against is XLM-R \cite{Conneau2020xlm-r}, the current state-of-the-art model for cross-lingual transfer \cite{Hu2020xtreme}. It is a Transformer-based model pretrained for 100 languages on large cleaned Common Crawl corpora \cite{Wenzek:2019arxiv}. For efficiency, we use the XLM-R Base configuration as the basis for most of our experiments. However, we note that the main idea behind the \model{} framework is not tied to any particular pretrained model, and the framework can be easily adapted to other pretrained multilingual models as we show later in \S\ref{sec:results} (e.g., multilingual BERT).
First, we compare against XLM-R in the standard setting where the entire model is fine-tuned on labelled data of the task in the source language.

\vspace{1.8mm}
\noindent \textbf{\mlmsrc{}; \mlmtrg{}}\hspace{0.3mm} 
In \S\ref{s:adaptation}, we have proposed target language adaptation as a simple method to adapt pretrained multilingual models for better cross-lingual generalisation on the downstream task. As a sanity check, we also compare against adapting to the source language data; we expect it to improve in-language performance but not to help with transfer. In particular, we fine-tune XLM-R with MLM on unlabelled source language (\mlmsrc{}) and target language data (\mlmtrg{}) prior to task-specific fine-tuning.

\subsection{ \model{}: Experimental Setup}

For the \textbf{\model{}} framework, unless noted otherwise, we rely on the XLM-R Base architecture; we evaluate the full \model{}, \model{} without invertible adapters (--\textsc{inv}), and also \model{} without language and invertible adapters (-- \textsc{lad} -- \textsc{inv}). We use the Transformers library \cite{Wolf2019transformers} for all our experiments. For fine-tuning via MLM on unlabelled data, we train on the Wikipedia data of the corresponding language for 250,000 steps, with a batch size of 64 and a learning rate of $5e-5$ and $1e-4$ for XLM-R (also for the \textsc{-src} and \textsc{-trg} variants) and adapters,  respectively. We train models on NER data for 100 epochs with a batch size of 16 and 8 for high-resource and low-resource languages, respectively, and a learning rate of $5e-5$ and $1e-4$ for XLM-R and adapters, respectively. We choose the best checkpoint for evaluation based on validation performance. Following \citet{Pfeiffer2020adapterfusion}, we learn language adapters, invertible adapters, and task adapters with dimensionalities of 384, 192 (384 for both directions), and 48, respectively. XLM-R Base has a hidden layer size of 768, so these adapter sizes correspond to reductions of 2, 2, and 16.

For NER, we conduct five runs of fine-tuning on the WikiAnn training set of the source language---except for \mlmtrg{} for which we conduct one run for efficiency purposes for every source language--target language combination. For QA, we conduct three runs of fine-tuning on the English SQuAD training set, evaluate on all XQuAD target languages, and report mean $F_1$ and exact match (EM) scores. For CCR, we conduct three runs of fine-tuning on the respective English training set, evaluate on all XCOPA target languages, and report accuracy scores.

\section{Results and Discussion} 
\label{sec:results}

\noindent \textbf{Named Entity Recognition}\hspace{0.3mm}
As our main summary of results, we average the cross-lingual transfer results of each method for each source language across all 16 target languages on the NER dataset. We show the aggregated results in Table~\ref{tab:main-results}. Moreover, in the appendix we report the detailed results for all methods across each single language pair, as well as a comparison of methods on the most common setting with English as source language.

In general, we observe that \xlmr{} performance is indeed lowest for unseen languages (the right half of the table after the vertical dashed line). \mlmsrc{} performs worse than \xlmr{}, which indicates that source-language fine-tuning is not useful for cross-lingual transfer in general.\footnote{However, there are some examples (e.g., \textsc{ja}, \textsc{tk}) where it does yield slight gains over the standard \xlmr{} transfer.} On the other hand, \mlmtrg{} is a stronger transfer method than \xlmr{} on average, yielding gains in 9/16 target languages. However, its gains seem to vanish for low-resource languages. Further, there is another disadvantage, outlined in \S\ref{s:adaptation}: \mlmtrg{} requires fine-tuning the full large pretrained model separately for each target language in consideration, which can be prohibitively expensive.

\model{} without language and invertible adapters performs on par with \xlmr{} for almost all languages present in the pretraining data (left half of the table). This mirrors findings in the monolingual setting where task adapters have been observed to achieve performance similar to regular fine-tuning while being more parameter-efficient \cite{Houlsby2019adapters}. However, looking at unseen languages, the performance of \model{} that only uses task adapters deteriorates significantly compared to \xlmr{}. This shows that task adapters alone are not expressive enough to bridge the discrepancy when adapting to an unseen language.

Adding language adapters to \model{} improves its performance across the board, and their usefulness is especially pronounced for low-resource languages. Language adapters help capture the characteristics of the target language and consequently provide boosts for unseen languages. Even for high-resource languages, the addition of language-specific parameters yields substantial improvements. Finally, invertible adapters provide further gains and generally outperform only using task and language adapters: for instance, we observe gains with \model{} over \model{} \textsc{--inv} on 13/16 target languages. Overall, the full \model{} framework improves upon \xlmr{} by more than 5 $F_1$ points on average.

To demonstrate that our framework is model-agnostic, we also employ two other strong multilingual models, XLM-R$^{Large}$ and mBERT as foundation for \model{} and show the results in Table \ref{tab:main-results}. \model{} shows consistent improvements even over stronger base pretrained models.

\begin{figure}[t!] 
\centering
\includegraphics[width=1.0\linewidth]{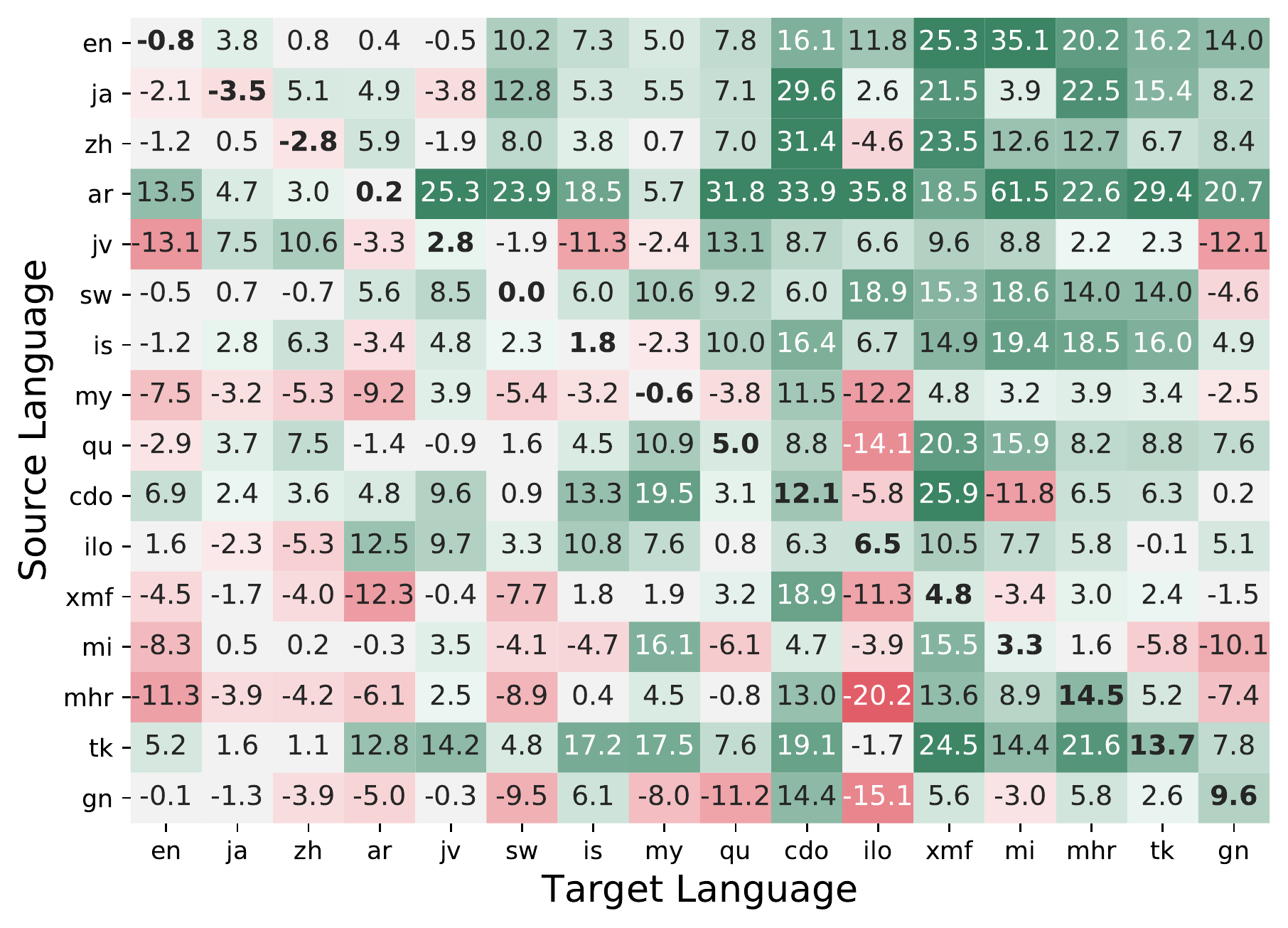}
\caption{Relative $F_1$ improvement of \model{}$^{Base}$ over \xlmr{}$^{Base}$ in cross-lingual NER transfer.}
\label{fig:invvsReg}
\end{figure}

For a more fine-grained impression of the performance of \model{} in different languages, we show its relative performance against \xlmr{} in the standard setting in Figure~\ref{fig:invvsReg}. We observe the largest differences in performance when transferring from high-resource to low-resource and unseen languages (top-right quadrant of Figure~\ref{fig:invvsReg}), which is arguably the most natural setup for cross-lingual transfer. In particular, we observe strong gains when transferring from Arabic, whose script might not be well represented in XLM-R's vocabulary. We also detect strong performance in the in-language monolingual setting (diagonal) for the subset of low-resource languages. This indicates that \model{} may help bridge the perceived weakness of multilingual versus monolingual models. Finally, \model{} performs competitively even when the target language is high-resource.\footnote{In the appendix, we also plot relative performance of the full \model{} method (with all three adapter types) versus \mlmtrg{} across all language pairs. The scores lead to similar conclusions as before: the largest benefits of \model{} are observed for the set of low-resource target languages (i.e., the right half of the heatmap). The scores also again confirm that the proposed \mlmtrg{} transfer baseline is more competitive than the standard \xlmr{} transfer across a substantial number of language pairs.}

\begin{table*}[]
\centering
\def\arraystretch{0.99}
{\footnotesize
\resizebox{\textwidth}{!}{%
\begin{tabular}{ll:lllllllllll | l}
\toprule
Model	&	en	&	et	&	ht	&	id	&	it	&	qu	&	sw	&	ta	&	th	&	tr	&	vi	&	zh	&	avg	\\
\midrule
\xlmr{}$^{Base}$	&	66.8	&	58.0	&	51.4	&	65.0	&	60.2	&	51.2	&	52.0	&	58.4	&	62.0	&	56.6	&	65.6	&	\textbf{68.8}	&	59.7	\\
\mlmtrg{} &	66.8	&	59.4	&	50.0	&	\textbf{71.0}	&	61.6	&	46.0	& \textbf{58.8}	&	60.0	&	\textbf{63.2}	&	\textbf{62.2}	&	\textbf{67.6}	&	67.4	&	61.2	\\
\midrule
\model{}$^{Base}$ &	\textbf{68.3}	&	\textbf{61.3}	&	\textbf{53.7}	&	65.8	&	\textbf{63.0}	&	\textbf{52.5}	&	56.3	&	\textbf{61.9}	&	61.8	&	60.3	&	66.1	&	67.6	&	\textbf{61.5}	\\
\bottomrule
\end{tabular}%
}
}%
\caption{Accuracy scores of all models on the XCOPA test sets when transferring from English. Models are first fine-tuned on SIQA and then on the COPA training set.}
\label{tab:xcopa-results}
\end{table*}

\begin{table*}[]
\centering
\resizebox{\textwidth}{!}{%
\begin{tabular}{l c : c c c c c c c c c c | c}
 \toprule
 & en & ar & de & el & es & hi & ru & th & tr & vi & zh & avg \\ \midrule
\xlmr{}$^{Base}$ &   83.6 / 72.1 &  66.8 / 49.1 & \textbf{74.4} / \textbf{60.1} & 73.0 / \textbf{55.7} & 76.4 / \textbf{58.3} & 68.2 / 51.7 & \textbf{74.3} / \textbf{58.1} & 66.5 / \textbf{56.7} &  68.3 /  52.8 & 73.7 / 53.8 & 51.3 / 42.0 & 70.6 / 55.5 \\ 
\mlmtrg{} 	&	\textbf{84.7} / \textbf{72.6}	&	\textbf{67.0} / \textbf{49.2}	&	73.7 / 58.8	&	\textbf{73.2} / 55.7	&	\textbf{76.6} / \textbf{58.3}	&	\textbf{69.8} / \textbf{53.6}	&	\textbf{74.3} / 57.9	&	67.0 / 55.8	&	\textbf{68.6} / \textbf{53.0}	&	\textbf{75.5} / \textbf{54.9}	&	52.2 / \textbf{43.1}	&	\textbf{71.1} / \textbf{55.7} \\
\midrule
\model{}$^{Base}$ $\:$ -- \textsc{inv} &	83.3 / 72.1	&	64.0 / 47.1	&	72.0 / 55.8	&	71.0 / 52.9	&	74.6 / 55.5	&	67.3 / 51.0	&	72.1 / 55.1	&	64.1 / 51.8	&	66.2 / 49.6	&	73.0 / 53.6	&	50.9 / 40.6	&	67.0 / 53.2 \\
\model{}$^{Base}$ &	83.5 / \textbf{72.6}	&	65.5 / 48.2	&	72.9 / 56.0	&	72.9 / 54.6	&	75.9 / 56.9	&	68.2 / 51.3	&	73.1 / 56.7	&	\textbf{67.8} / 55.9	&	67.0 / 49.8	&	 73.7 / 53.3	&	\textbf{52.7} / 42.8	&	70.3 / 54.4\\
\bottomrule
\end{tabular}%
}
\caption{$F_1$ / EM scores on XQuAD with English as the source language for each target language.}
\label{tab:xquad-results}
\end{table*}

\vspace{1.8mm}
\noindent \textbf{Causal Commonsense Reasoning}\hspace{0.3mm} \label{ss:ccr}
We show results on transferring from English to each target language on XCOPA in Table~\ref{tab:xcopa-results}. For brevity, we only show the results of the best fine-tuning setting from \newcite{Ponti2020xcopa}---fine-tuning first on SIQA \cite{sap2019socialiqa} and on the English COPA training set---and report other possible settings in the appendix. Target language adaptation outperforms \xlmr{}$^{Base}$ while \model{}$^{Base}$ achieves the best scores. It shows gains in particular for the two unseen languages, Haitian Creole (ht) and Quechua (qu). Performance on the other languages is also generally competitive or better.

\vspace{1.8mm}
\noindent \textbf{Question Answering}\hspace{0.3mm} \label{ss:qa}
The results on XQuAD when transferring from English to each target language are provided in Table \ref{tab:xquad-results}. The main finding is that \model{} achieves similar performance to the \xlmr{} baseline. As before, invertible adapters generally improve performance and target language adaptation improves upon the baseline setting. We note that all languages included in XQuAD can be considered high-resource, with more than 100k Wikipedia articles each (cf. Wikipedia sizes of NER languages in Table~\ref{tab:ner-languages}). The corresponding setting can be found in the top-left quadrant in Figure~\ref{fig:invvsReg} where relative differences are comparable.

These and XCOPA results demonstrate that, while \model{} excels at transfer to unseen and low-resource languages, it achieves competitive performance even for high-resource languages and on more challenging tasks. These evaluations also hint at the modularity of the adapter-based \model{} approach, which holds promise of quick adaptation to more tasks: we use exactly the same language-specific adapters in NER, CCR, and QA for languages such as English and Mandarin Chinese that appear in all three evaluation language samples.

\section{Further Analysis}

\noindent \textbf{Impact of Invertible Adapters}\hspace{0.3mm} 
We also analyse the relative performance difference of \model{} with and without invertible adapters for each source language--target language pair on the NER data set (see Section D in the appendix). 
Invertible adapters improve performance for many transfer pairs, and particularly when transferring to low-resource languages. Performance is only consistently lower with a single low-resource language as source (Maori), likely due to variation in the data.

 \begin{figure} 
\centering
\includegraphics[width=1.0\linewidth]{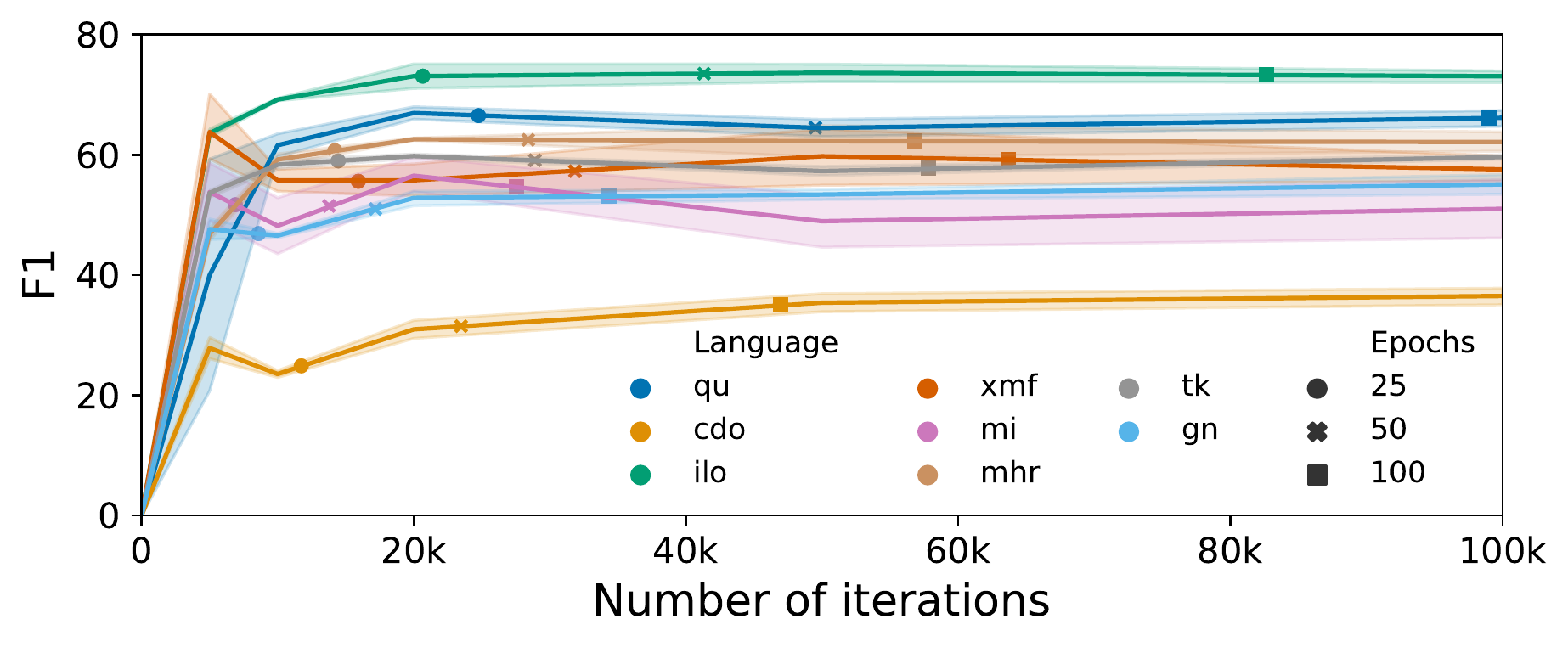}
\caption{Cross-lingual NER performance of \model{} transferring from English to the target languages with invertible and language adapters trained on target language data for different numbers of iterations. Shaded regions denote variance in $F_1$ scores across 5 runs.}
\label{fig:sample_eff}
\end{figure}

\begin{table}[]
\centering
\def\arraystretch{0.99}
{\footnotesize
\resizebox{\columnwidth}{!}
{%
\begin{tabular}{l cc}
\toprule
Model	&	+ Params	&	\% Model	  \\
\midrule
\model{}$^{Base}$	 &	 8.25M	&  3.05  	\\
\model{}$^{Base}$ $\:$ -- \textsc{inv}	& 7.96M	& 2.94	\\
\model{}$^{Base}$ $\:$ -- \textsc{lad} -- \textsc{inv}	&	0.88M 	&	0.32 	\\
\bottomrule
\end{tabular}%
}
} 
\caption{Number of parameters added to  
XLM-R Base, 
and as a fraction of its parameter budget (270M).}
\label{tab:parameters} 
\end{table}

\vspace{1.8mm}
\noindent \textbf{Sample Efficiency}\hspace{0.3mm}
The main adaptation bottleneck of \model{} is training language adapters and invertible adapters. However, due to the modularity of \model{}, once trained, these adapters have an advantage of being directly reusable (i.e., ``plug-and-playable'') across different tasks (see the discussion in \S\ref{ss:qa}). To estimate the sample efficiency of adapter training, we measure NER performance on several low-resource target languages (when transferring from English as the source) conditioned on the number of training iterations. The results are given in Figure~\ref{fig:sample_eff}. They reveal that we can achieve strong performance for the low-resource languages already at 20k training iterations, and longer training offers modest increase in performance.

Moreover, in Table~\ref{tab:parameters} we present the number of parameters added to the original XLM-R Base model per language for each \model{} variant. The full \model{} model for NER receives an additional set of 8.25M adapter parameters for every language, which makes up only 3.05\% of the original model.

\section{Conclusion}

We have proposed \model{}, a general modular framework for transfer across tasks and languages. It leverages a small number of additional parameters to mitigate the capacity issue which fundamentally hinders current multilingual models. \model{} is model-agnostic and can be adapted to any current pre-trained multilingual model as foundation. We have shown that it is particularly useful for adapting to languages not covered by the multilingual model's training model, while also achieving competitive performance on high-resource languages.

In future work, we will apply \model{} to other pre-trained models, and employ adapters that are particularly suited for languages with certain properties (e.g. with different scripts). We will also evaluate on additional tasks, and investigate leveraging pre-trained language adapters from related languages for improved transfer to truly low-resource languages with limited monolingual data.

\section*{Acknowledgments}
Jonas Pfeiffer is supported by the LOEWE initiative (Hesse, Germany) within the emergenCITY center. The work of Ivan Vuli\'{c} is supported by the ERC Consolidator Grant LEXICAL: Lexical Acquisition Across Languages (no 648909). We thank Laura Rimell for feedback on a draft.

We would like to thank \href{https://instagram.com/isabelpfeiffer_art?igshid=165k7u2wz7pb0}{Isabel Pfeiffer} for the logo illustrations.

\begin{textblock*}{5cm}(13.4cm,26.0cm)  
\includegraphics[width=5cm]{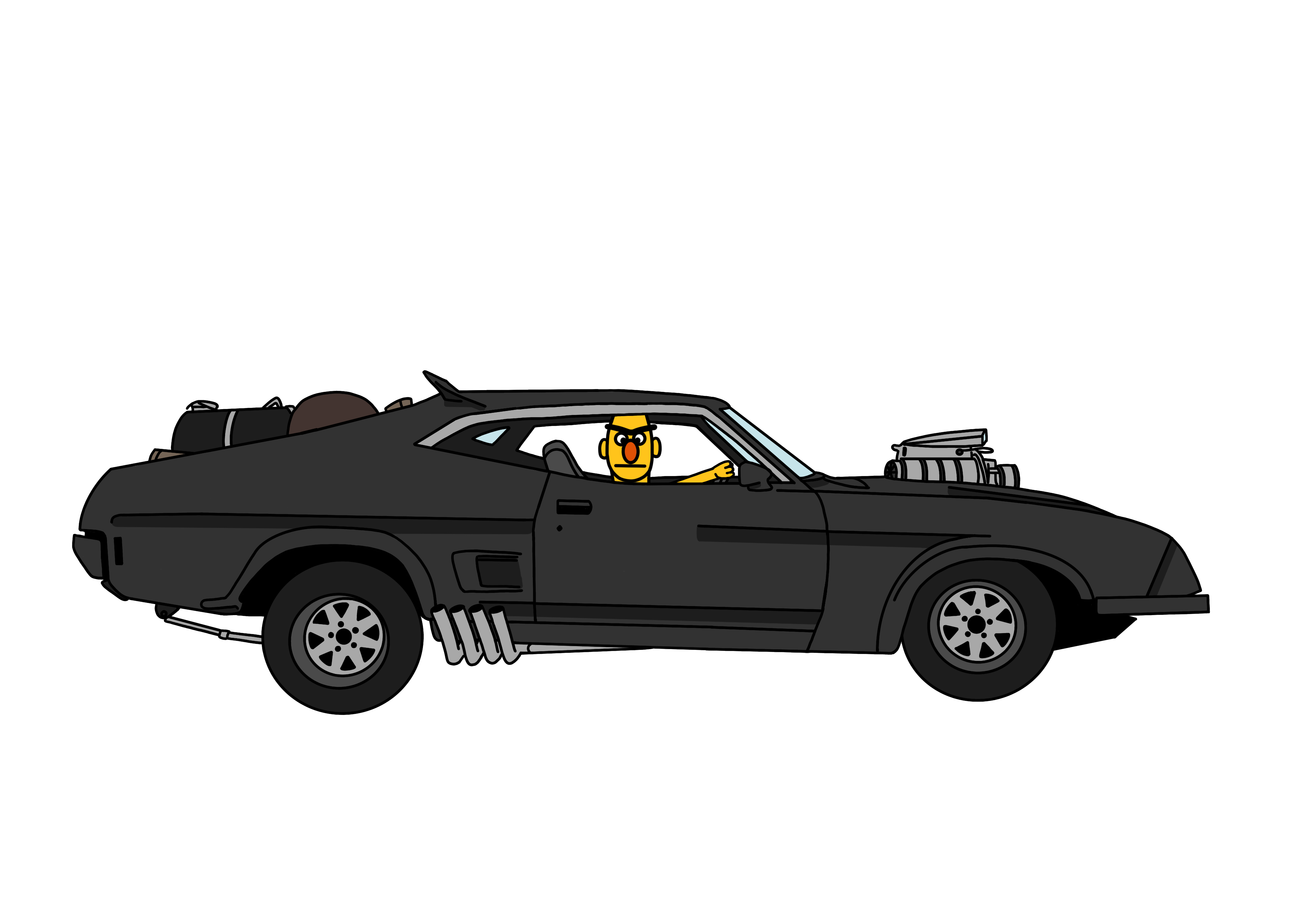}
\end{textblock*}

\bibliography{emnlp2020}
\bibliographystyle{acl_natbib}

\appendix
\appendix

\begin{table*}[!t]
\centering
\resizebox{\textwidth}{!}{%
\begin{tabular}{ll|lllllll:llllllll | l}
\toprule
	&	en	&	ja	&	zh	&	ar	&	jv	&	sw	&	is	&	my	&	qu	&	cdo	&	ilo	&	xmf	&	mi	&	mhr	&	tk	&	gn	& avg \\
	\midrule
mBERT 	&	84.8	&	\textbf{26.7}	&	\textbf{38.5}	&	38.7	&	\textbf{57.8}	&	66.0	&	65.7	&	42.9	&	54.9	&	14.20	&	63.5	&	31.1	&	21.8	&	46.0	&	47.2	&	45.4	&	44.0 \\
XLM-R 	&	83.0	&	15.2	&	19.6	&	41.3	&	56.1	&	63.5	&	67.2	&	46.9	&	58.3	&	20.47	&	61.3	&	32.2	&	15.9	&	41.8	&	43.4	&	41.0	&	41.6 \\
\mlmsrc{}	&	84.2	&	8.45	&	11.0	&	27.3	&	44.8	&	57.9	&	59.0	&	35.6	&	52.5	&	21.4	&	60.3	&	22.7	&	22.7	&	38.1	&	44.0	&	41.7	&	36.5 \\
\mlmtrg{}	&	84.2	&	9.30	&	15.5	&	\textbf{44.5}	&	50.2	&	\textbf{77.7}	&	71.7	&	\textbf{55.5}	&	\textbf{68.7}	&	\textbf{47.6}	&	\textbf{84.7}	&	\textbf{60.3}	&	43.6	&	56.3	&	56.4	&	50.6	&	52.8 \\
\midrule
\model{} $\:$ -- \textsc{lad} -- \texttt{inv}	&   82.0	&	15.6	&	20.3	&	41.0	&	54.4	&	66.4	&	67.8	&	48.8	&	57.8	&	16.9	&	59.9	&	36.9	&	14.3	&	44.3	&	41.9	&	42.9	&	41.9 \\ 
\model{} $\:$ -- \textsc{inv}	&	82.2	&	16.8	&	20.7	&	36.9	&	54.1	&	68.7	&	71.5	&	50.0	&	59.6	&	39.2	&	69.9	&	54.9	&	48.3	&	58.1	&	53.1	&	52.8	&	50.3 \\
\model{} $\:$ 	&	82.3	&	19.0	&	20.5	&	41.8	&	55.7	&	73.8	&	\textbf{74.5}	&	51.9	&	66.1	&	36.5	&	73.1	&	57.6	&	\textbf{51.0}	&	\textbf{62.1}	&	\textbf{59.7}	&	\textbf{55.1}	&	\textbf{53.2} \\
\bottomrule
\end{tabular}%
}
\caption{NER F1 scores for zero-shot transfer from English.}
\label{tab:main-results-en}
\end{table*}

\section{Evaluation data}
\label{task:data}
\begin{itemize}
    \item Named Entity Recognition (NER). Data: WikiANN \cite{Rahimi2019massively}. Available online at: \\ \url{www.amazon.com/clouddrive/share/d3KGCRCIYwhKJF0H3eWA26hjg2ZCRhjpEQtDL70FSBN}.
    \item Causal Commonsense Reasoning (CCR). Data: XCOPA \cite{Ponti2020xcopa}. Available online at: \\
    \url{github.com/cambridgeltl/xcopa}
    \item Question Answering (QA). Data: XQuAD \cite{Artetxe2020cross-lingual}. Available online at: \\
    \url{github.com/deepmind/xquad}
    
\end{itemize}

\section{NER zero-shot results from English}
\label{s:ner-zs-en}

We show the F1 scores when transferring from English to the other languages averaged over five runs in Table~\ref{tab:main-results-en}.

\section{NER results per language pair}
\label{s:ner-all-pairs}

\begin{figure*} 
\centering
\includegraphics[width=0.9\linewidth]{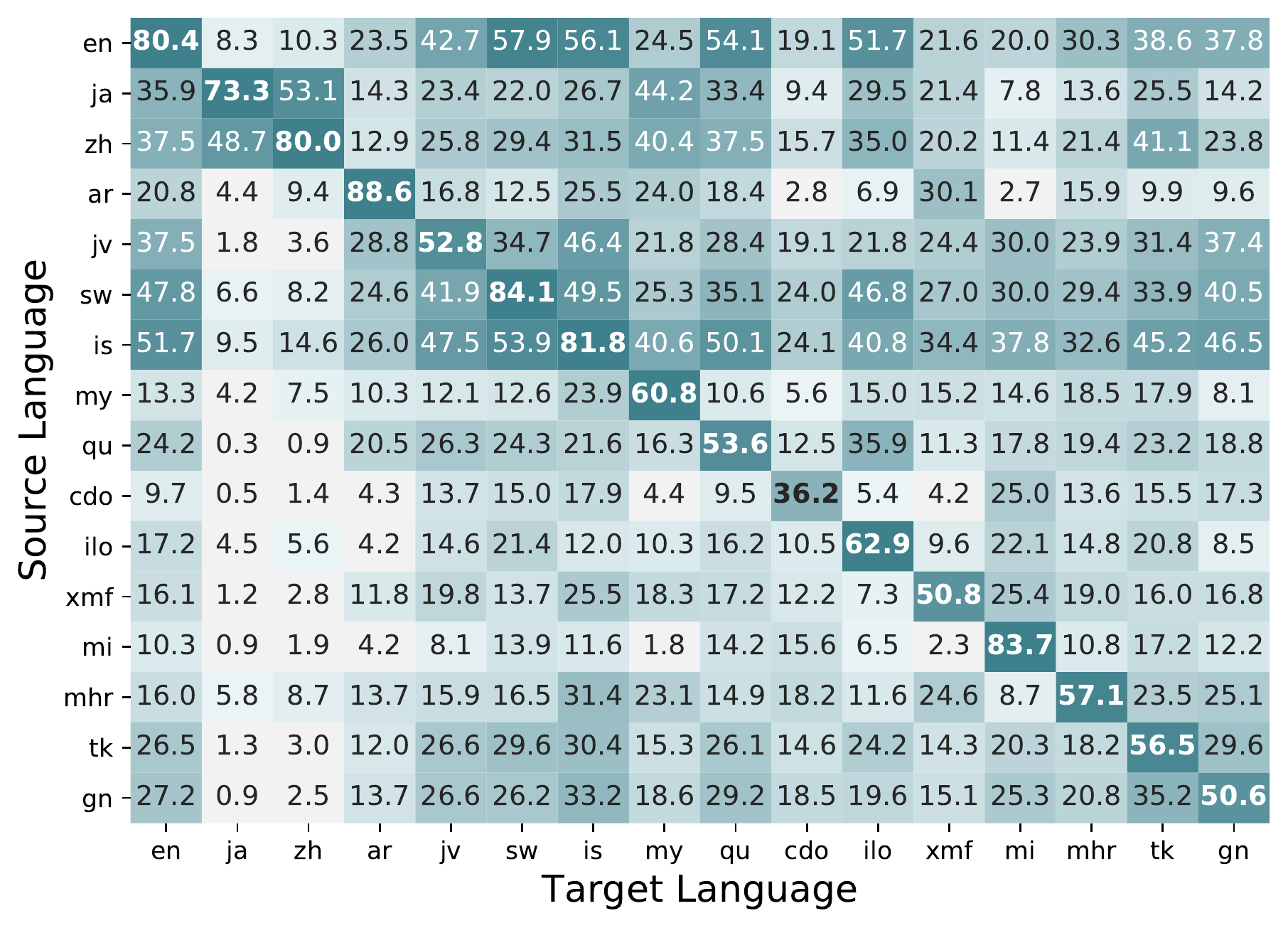}
\caption{Mean F1 scores of XLM-R$^{Base}$ in the standard setting (\xlmr{}$^{Base}$) for cross-lingual transfer on NER.}
\label{fig:reg_scores_hm}
\end{figure*}

\begin{figure*} 
\centering
\includegraphics[width=0.9\linewidth]{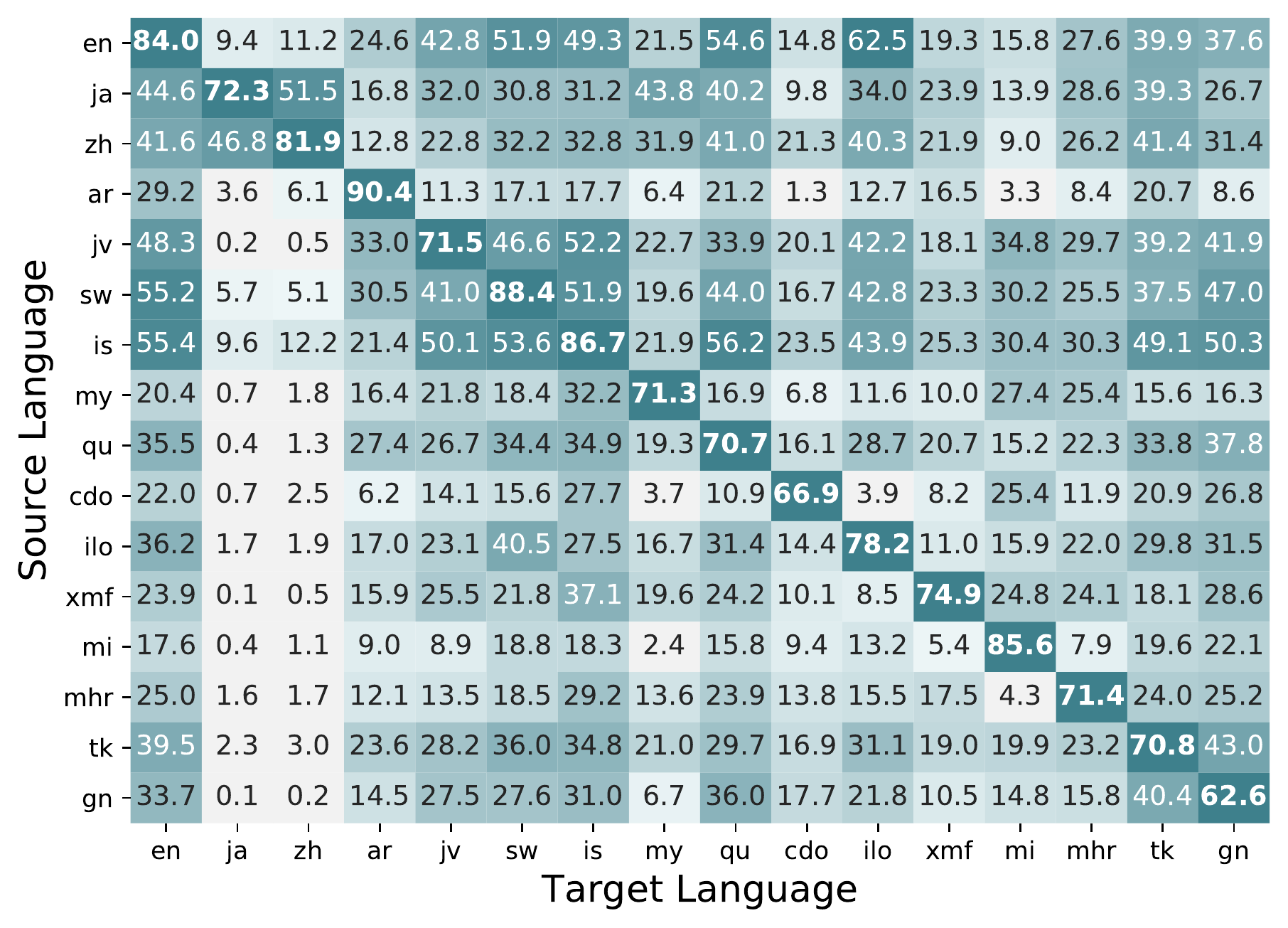}
\caption{Mean F1 scores of XLM-R$^{Base}$ with MLM fine-tuning on source language data (\mlmsrc{}) for cross-lingual transfer on NER.}
\label{fig:full_scores_hm}
\end{figure*}

\begin{figure*} 
\centering
\includegraphics[width=0.9\linewidth]{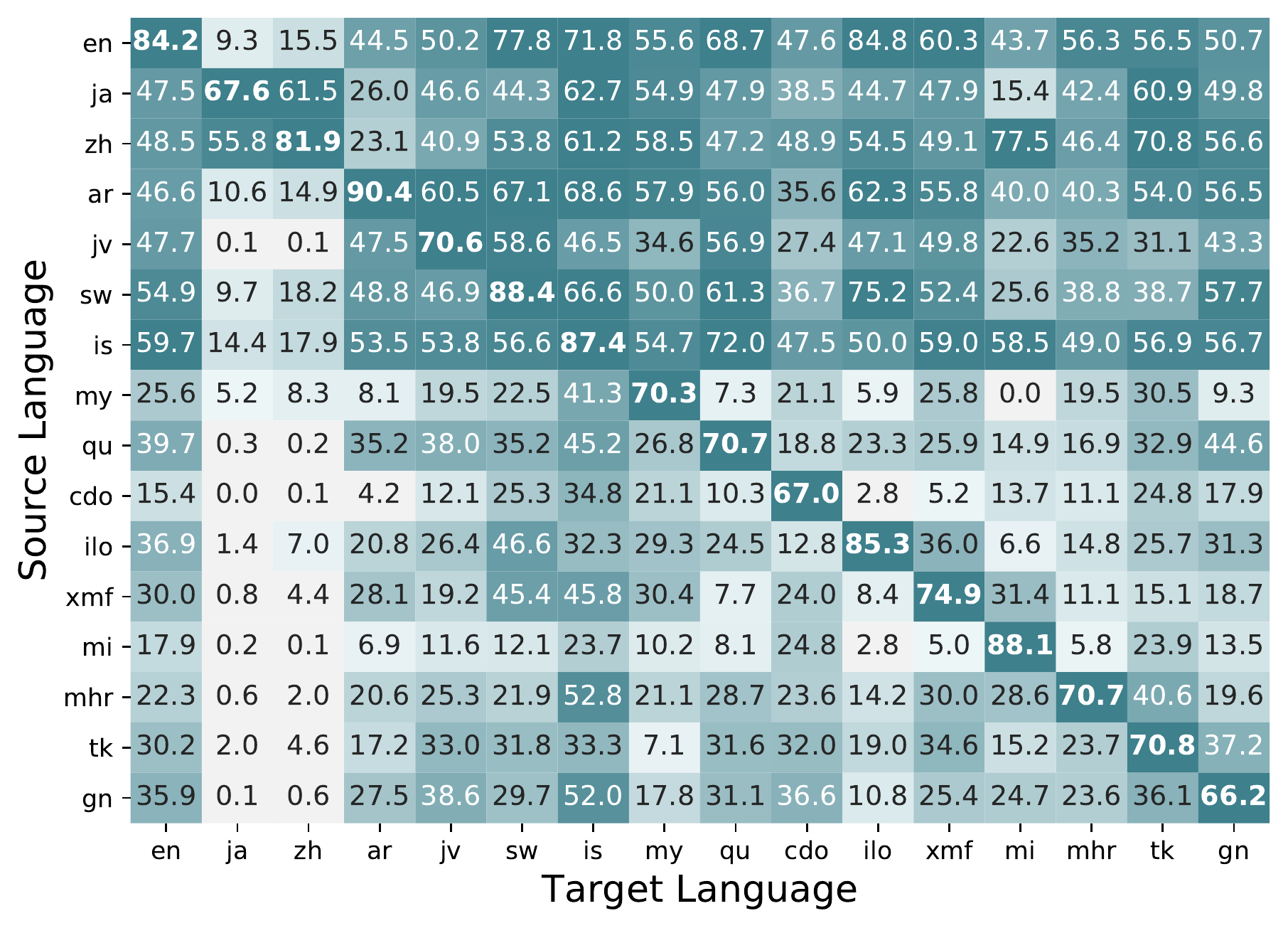}
\caption{Mean F1 scores of XLM-R$^{Base}$ with MLM fine-tuning on target language data (\mlmtrg{}) for cross-lingual transfer on NER.}
\label{fig:trg_scores_hm}
\end{figure*}

\begin{figure*} 
\centering
\includegraphics[width=0.9\linewidth]{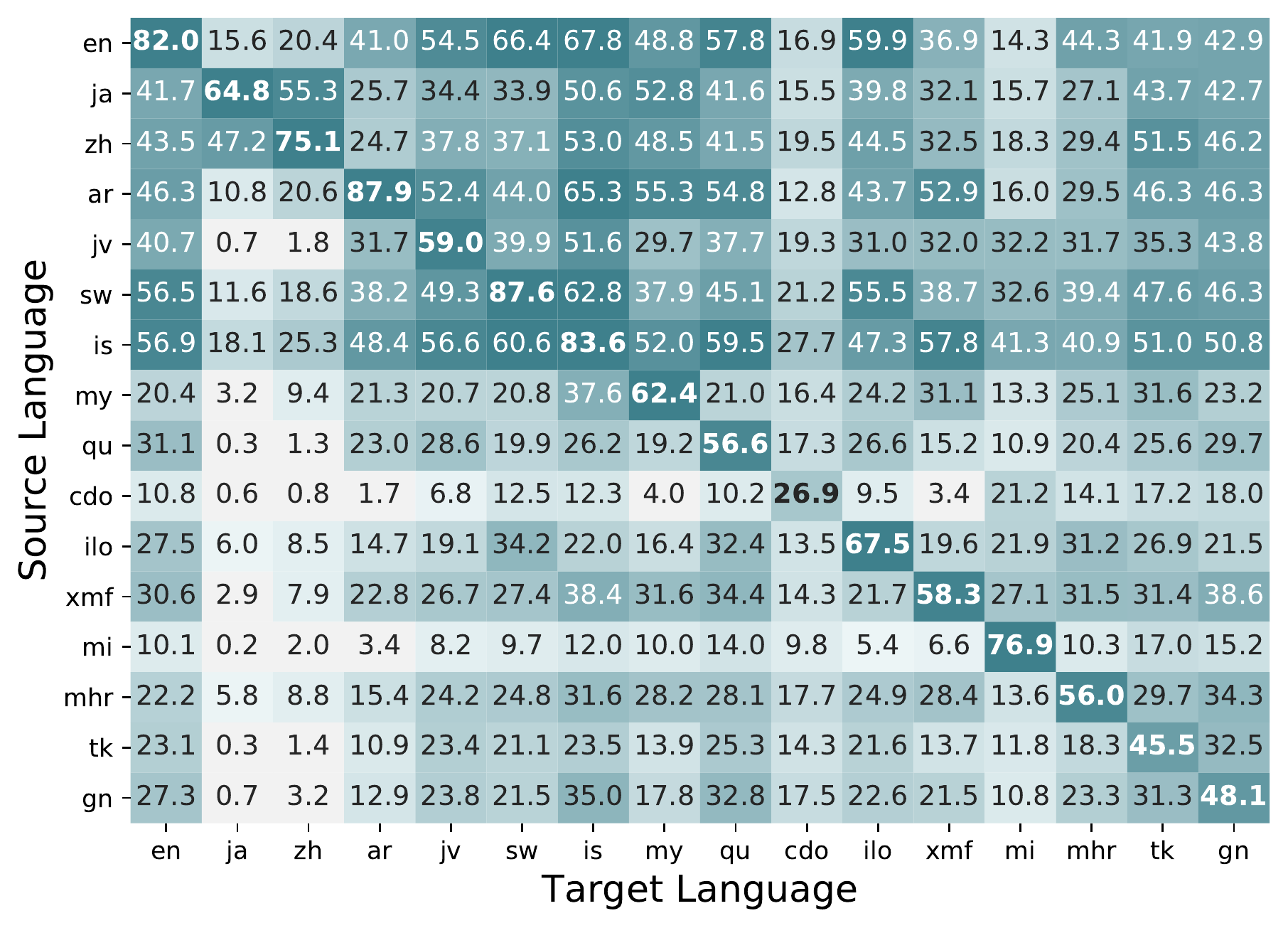}
\caption{Mean F1 scores of our framework without language adapters and invertible adapters (\model{}$^{Base}$ -- \textsc{lad} -- \textsc{inv}) for cross-lingual transfer on NER.}
\label{fig:nolad_scores_hm}
\end{figure*}

\begin{figure*} 
\centering
\includegraphics[width=0.9\linewidth]{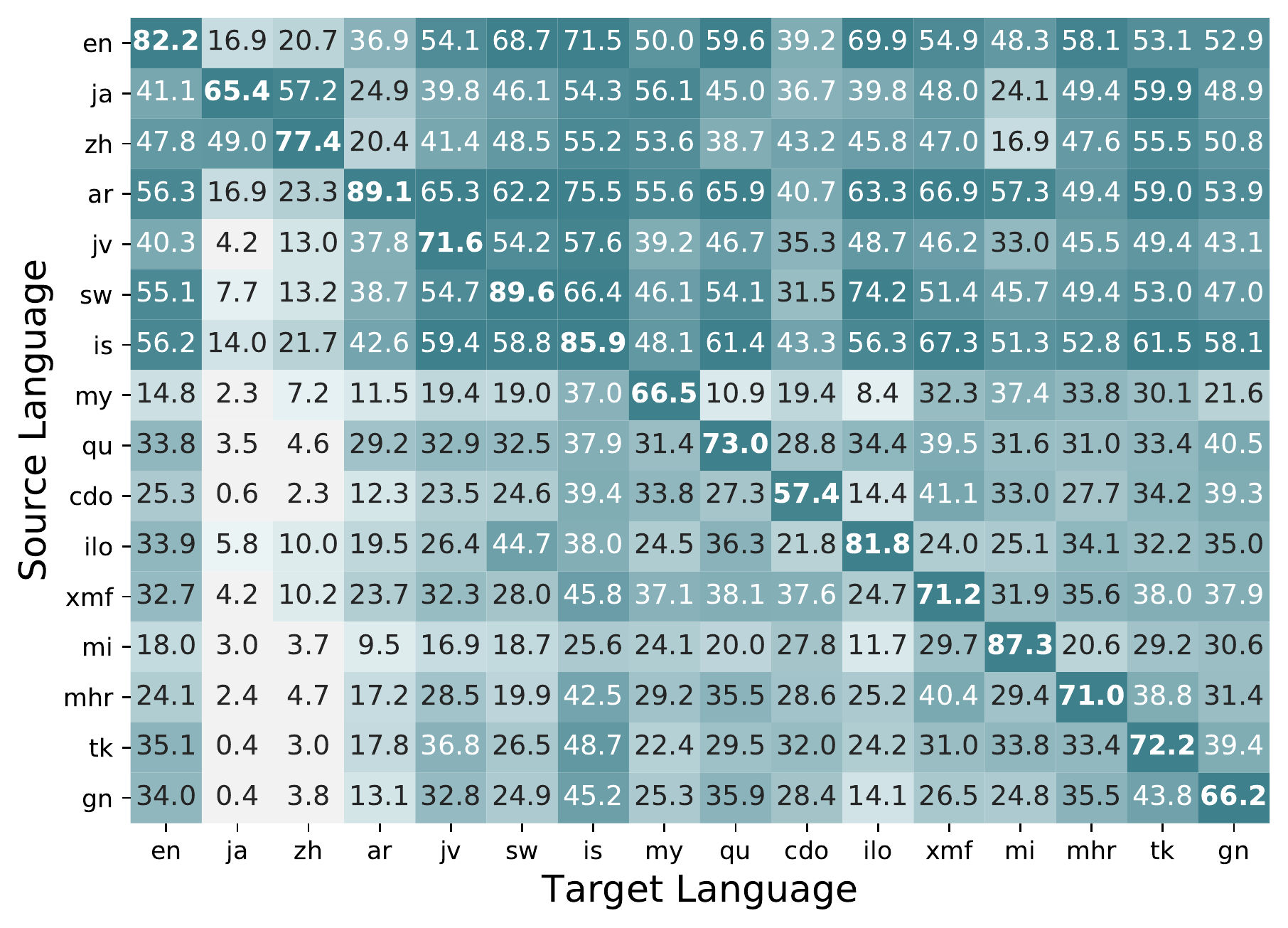}
\caption{Mean F1 scores of our framework without invertible adapters (\model{}$^{Base}$ -- \textsc{inv}) for cross-lingual transfer on NER.}
\label{fig:adapters_scores_hm}
\end{figure*}

\begin{figure*} 
\centering
\includegraphics[width=0.9\linewidth]{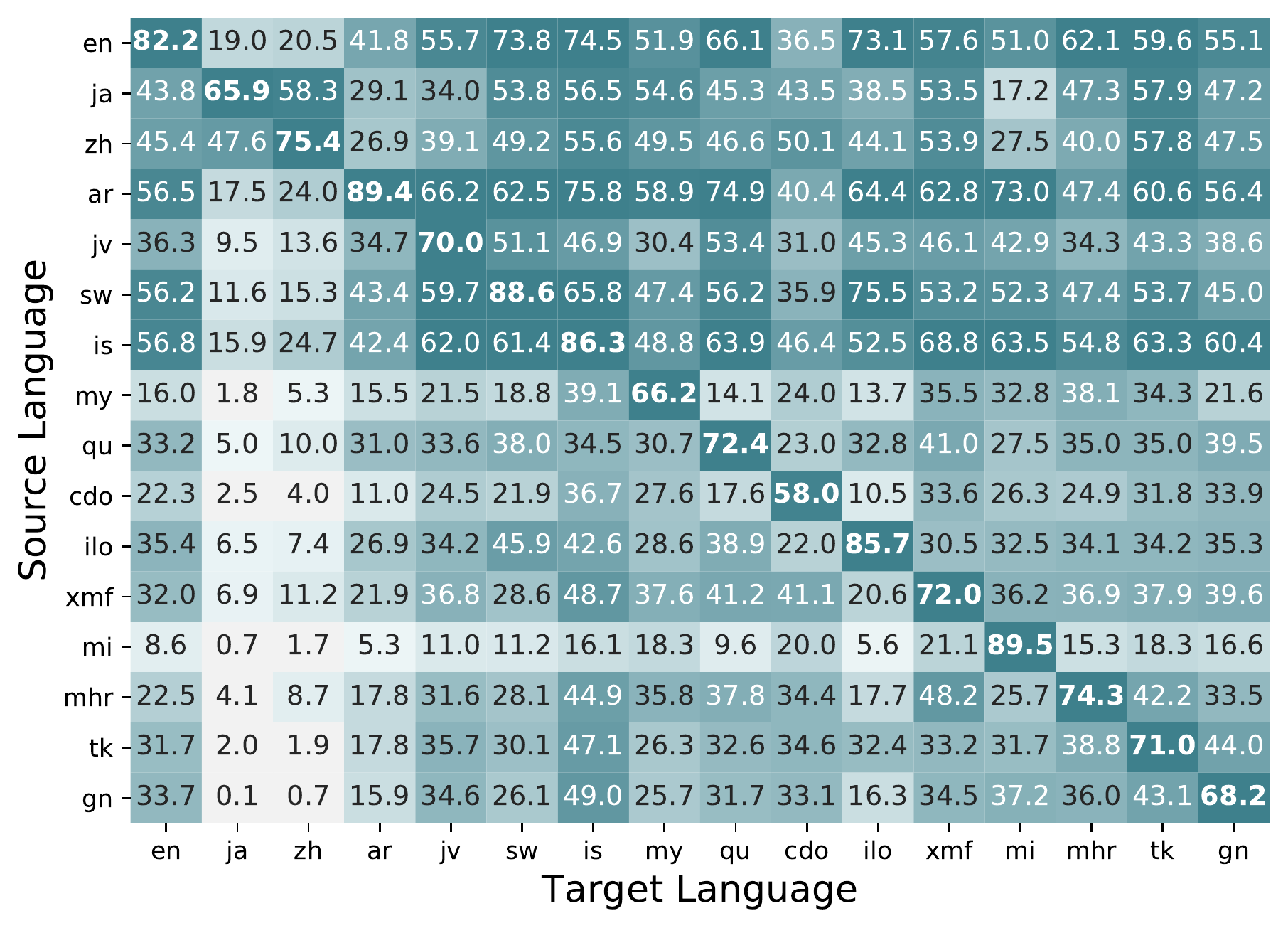}
\caption{Mean F1 scores of our complete adapter-based framework (\model{}$^{Base}$) for cross-lingual transfer on NER.}
\label{fig:inv_scores_hm}
\end{figure*}

\begin{figure*} 
\centering
\includegraphics[width=0.9\linewidth]{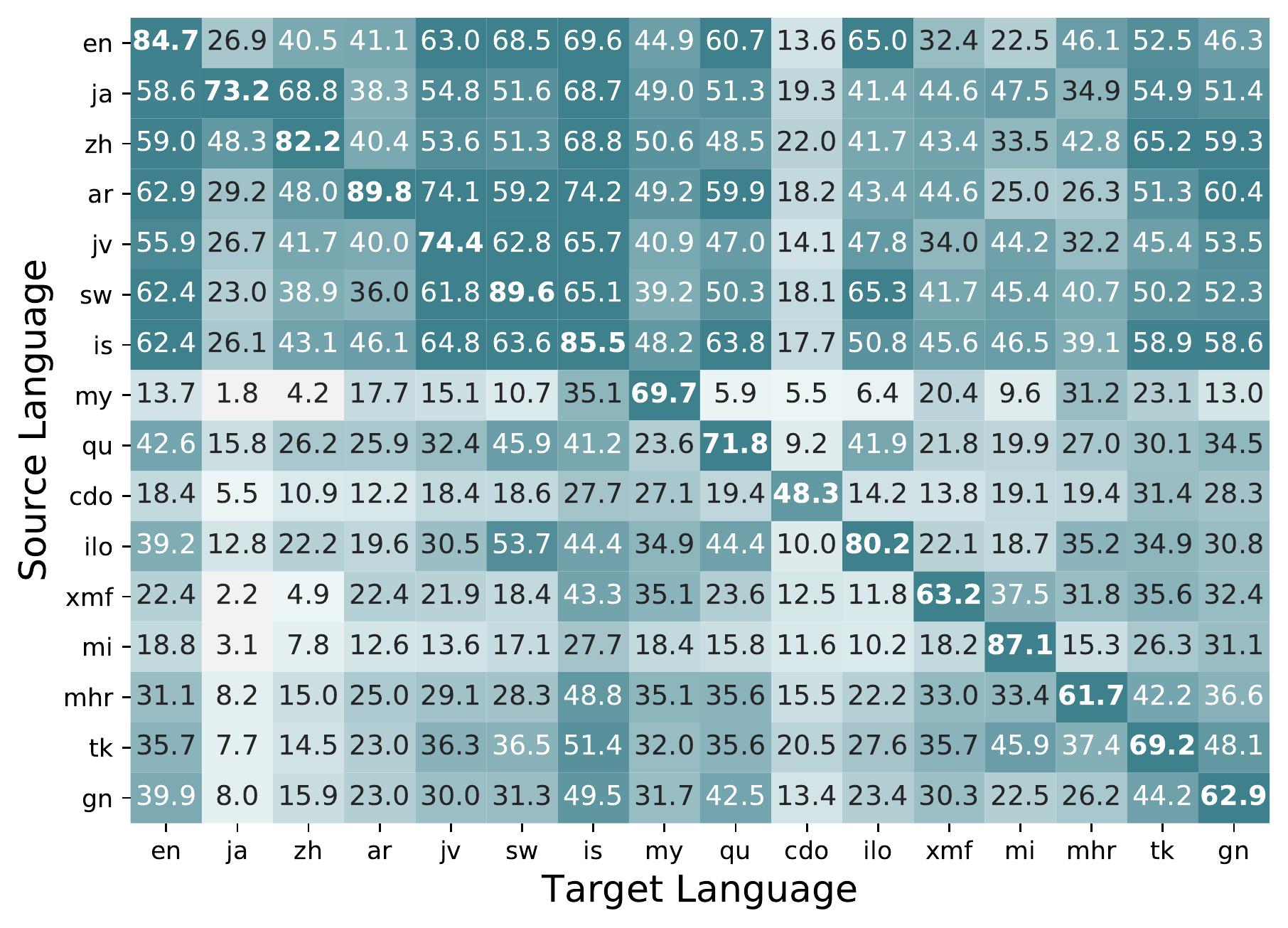}
\caption{Mean F1 scores of mBERT for cross-lingual transfer on NER.}
\label{fig:mbert_scores}
\end{figure*}

\begin{figure*} 
\centering
\includegraphics[width=0.9\linewidth]{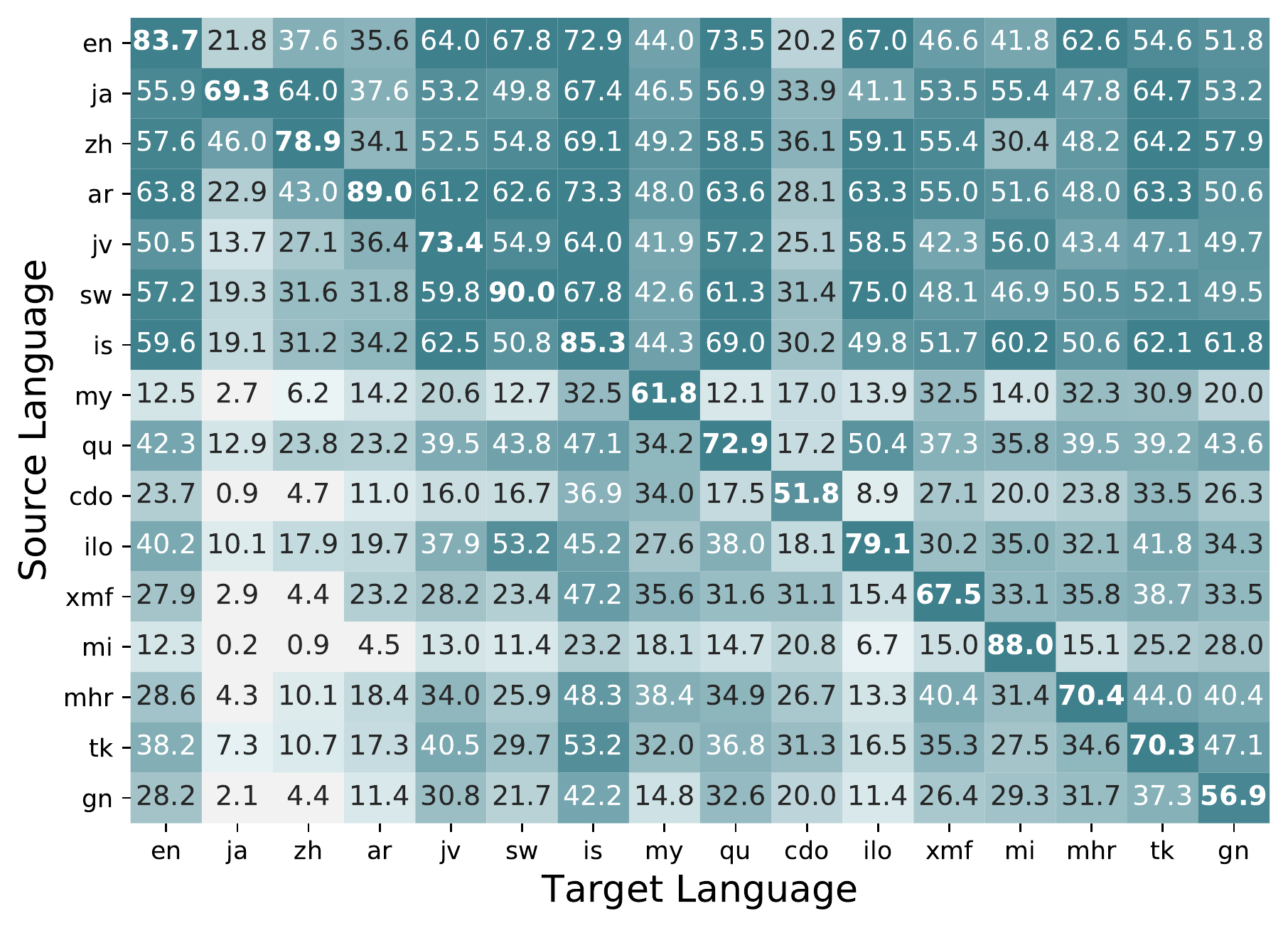}
\caption{Mean F1 scores of our complete adapter-based framework (\model{}$^{mBERT}$) for cross-lingual transfer on NER.}
\label{fig:mbert_adap_scores}
\end{figure*}

\begin{figure*} 
\centering
\includegraphics[width=0.9\linewidth]{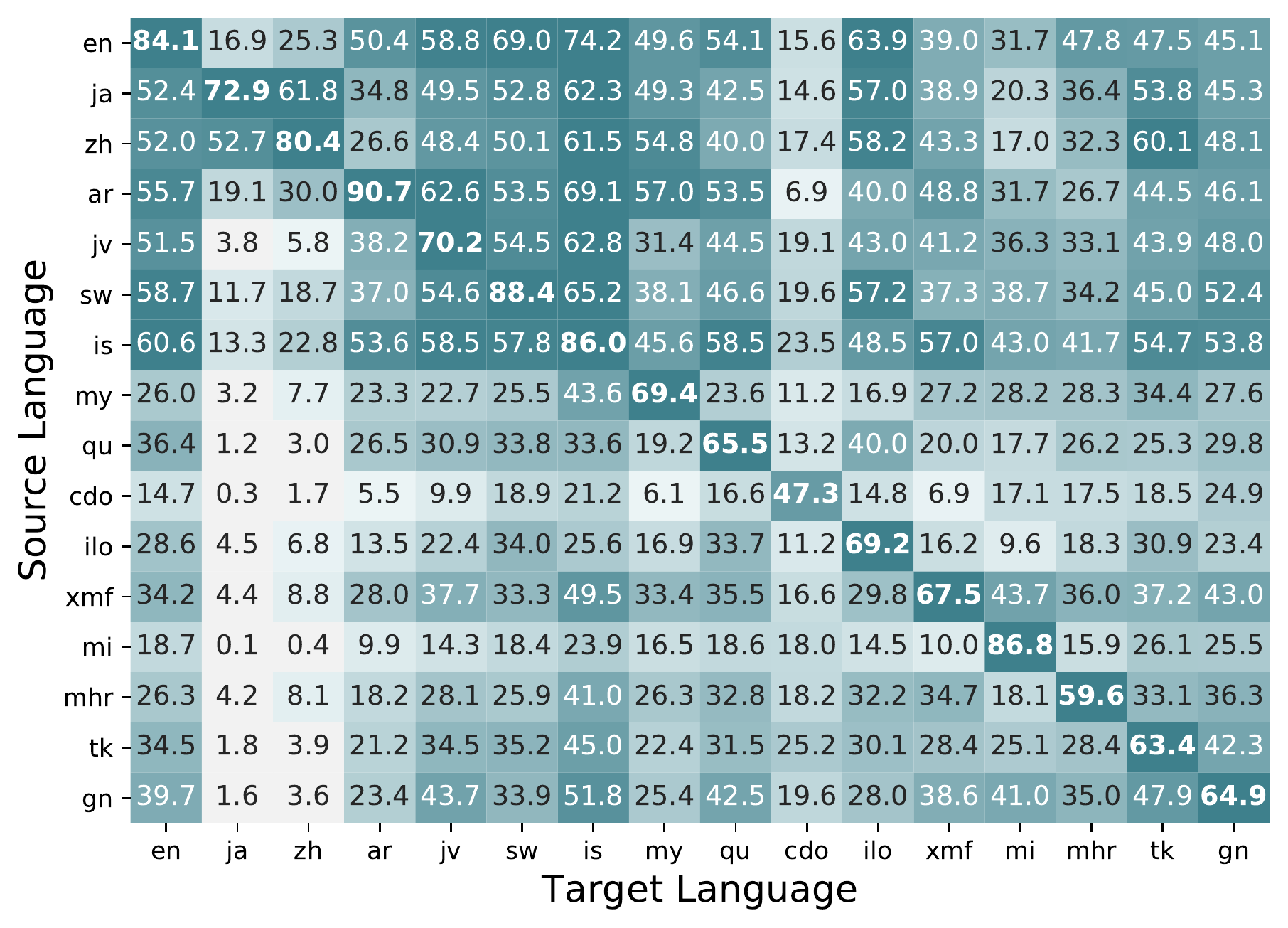}
\caption{Mean F1 scores of XLM-R$^{Large}$ for cross-lingual transfer on NER.}
\label{fig:large_scores}
\end{figure*}

\begin{figure*} 
\centering
\includegraphics[width=0.9\linewidth]{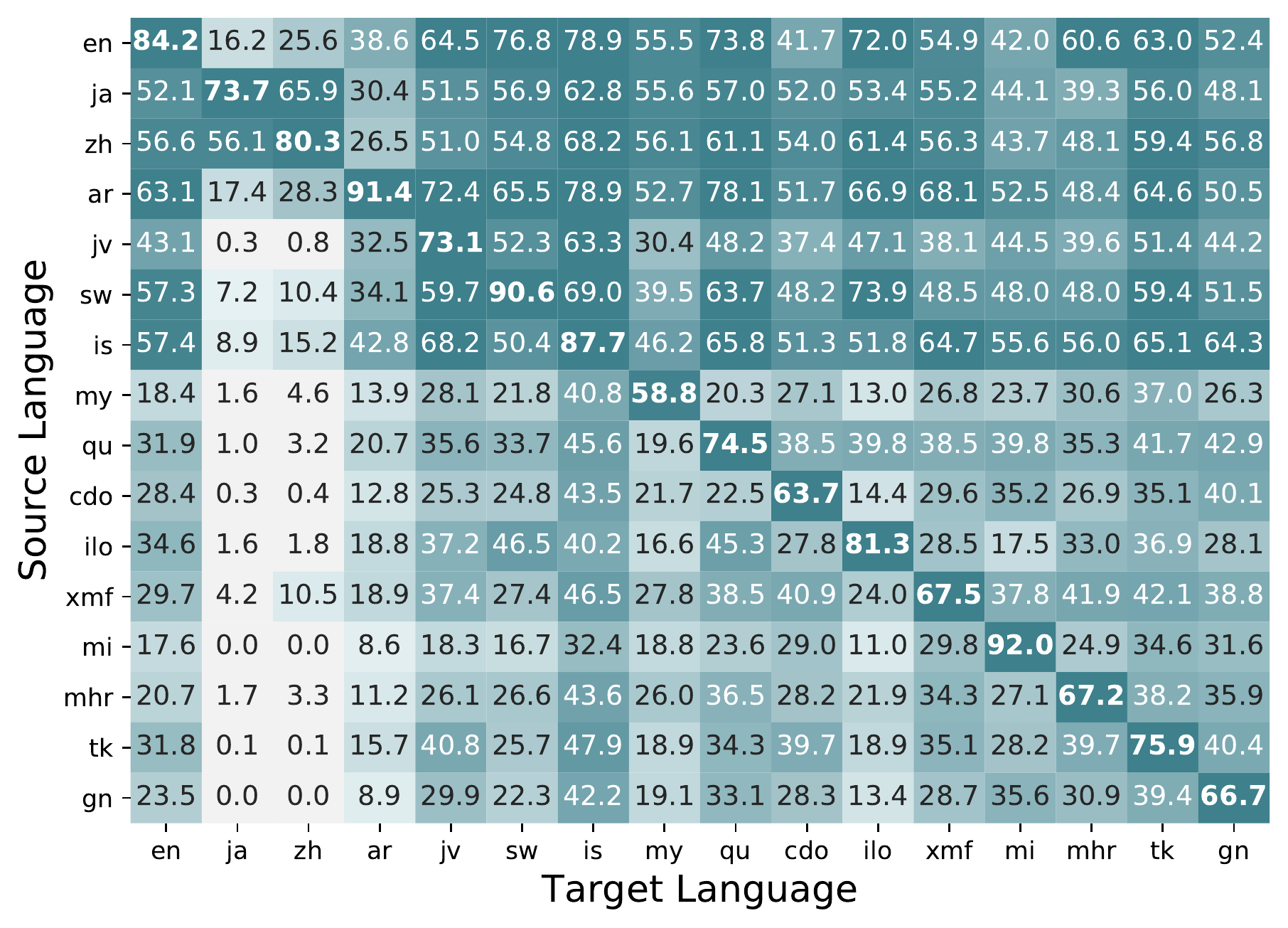}
\caption{Mean F1 scores of our complete adapter-based framework (\model{}$^{Large}$) for cross-lingual transfer on NER.}
\label{fig:large_adap_scores}
\end{figure*}

We show the F1 scores on the NER dataset across all combinations of source and target language for all of our comparison methods in Figures \ref{fig:reg_scores_hm} (\xlmr{}$^{Base}$), \ref{fig:full_scores_hm} (\mlmsrc{}), \ref{fig:trg_scores_hm} (\mlmtrg{}), \ref{fig:nolad_scores_hm} (\model{}$^{Base}$ -- \textsc{lad} -- \textsc{inv}), \ref{fig:adapters_scores_hm} (\model{}$^{Base}$ -- \textsc{inv}), \ref{fig:inv_scores_hm} (\model{}$^{Base}$), \ref{fig:mbert_scores} (mBERT), \ref{fig:mbert_adap_scores} (\model{}$^{mBERT}$) , \ref{fig:large_scores} (\xlmr{}$^{Large}$), and  \ref{fig:large_adap_scores} (\model{}$^{mBERT}$). Each score is averaged over five runs. 

\section{Relative improvement of MAD-X over baselines in cross-lingual NER transfer}
\label{s:relative-appendix}
\begin{figure*}[b] 
\centering
\includegraphics[width=0.9\textwidth]{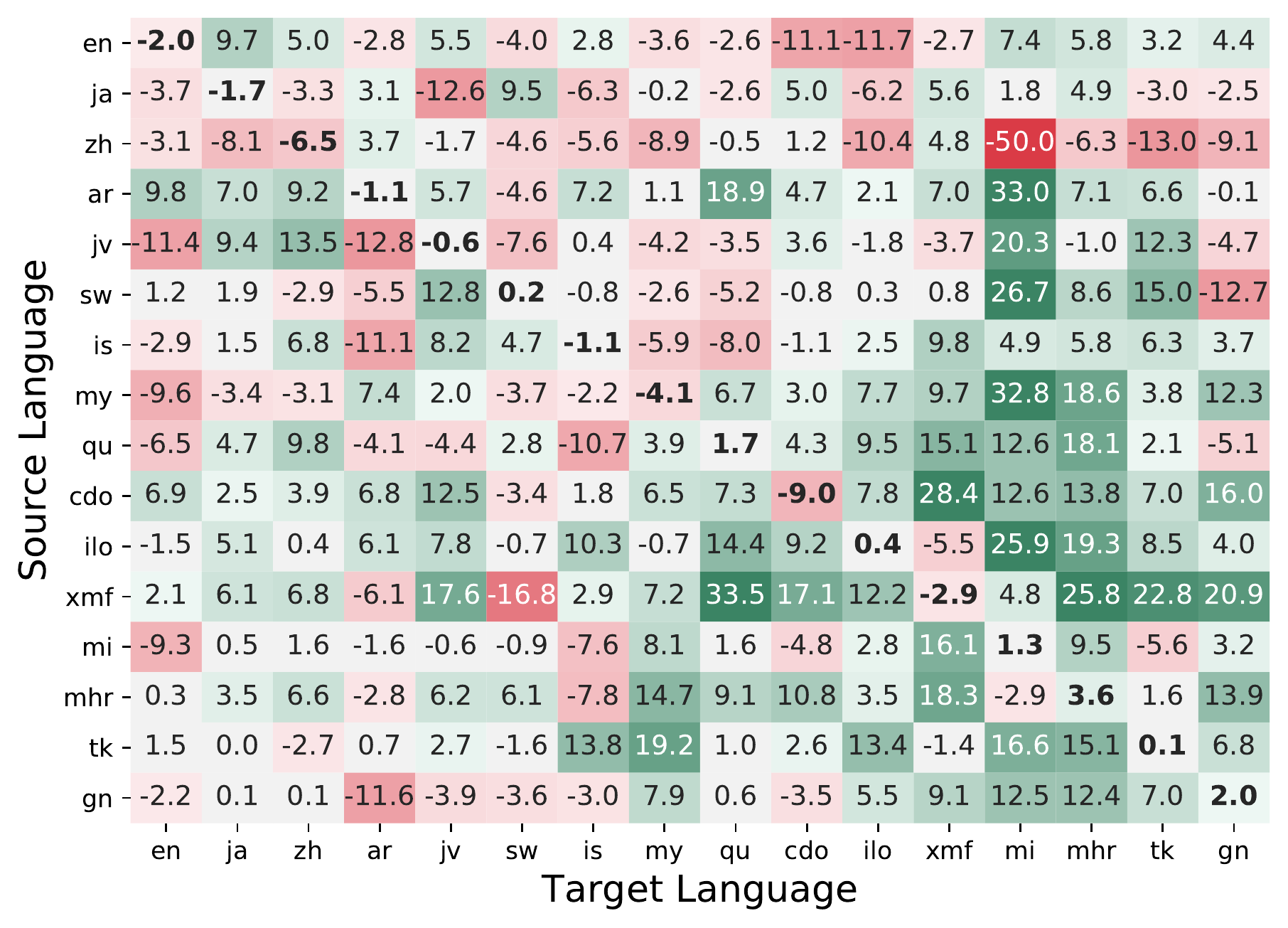}
\caption{Relative $F_1$ improvement of \model{}$^{Base}$ over \mlmtrg{} in cross-lingual NER transfer.}
\label{fig:invvstrg}
\end{figure*}

\begin{figure*}[b] 
\centering
\includegraphics[width=0.9\textwidth]{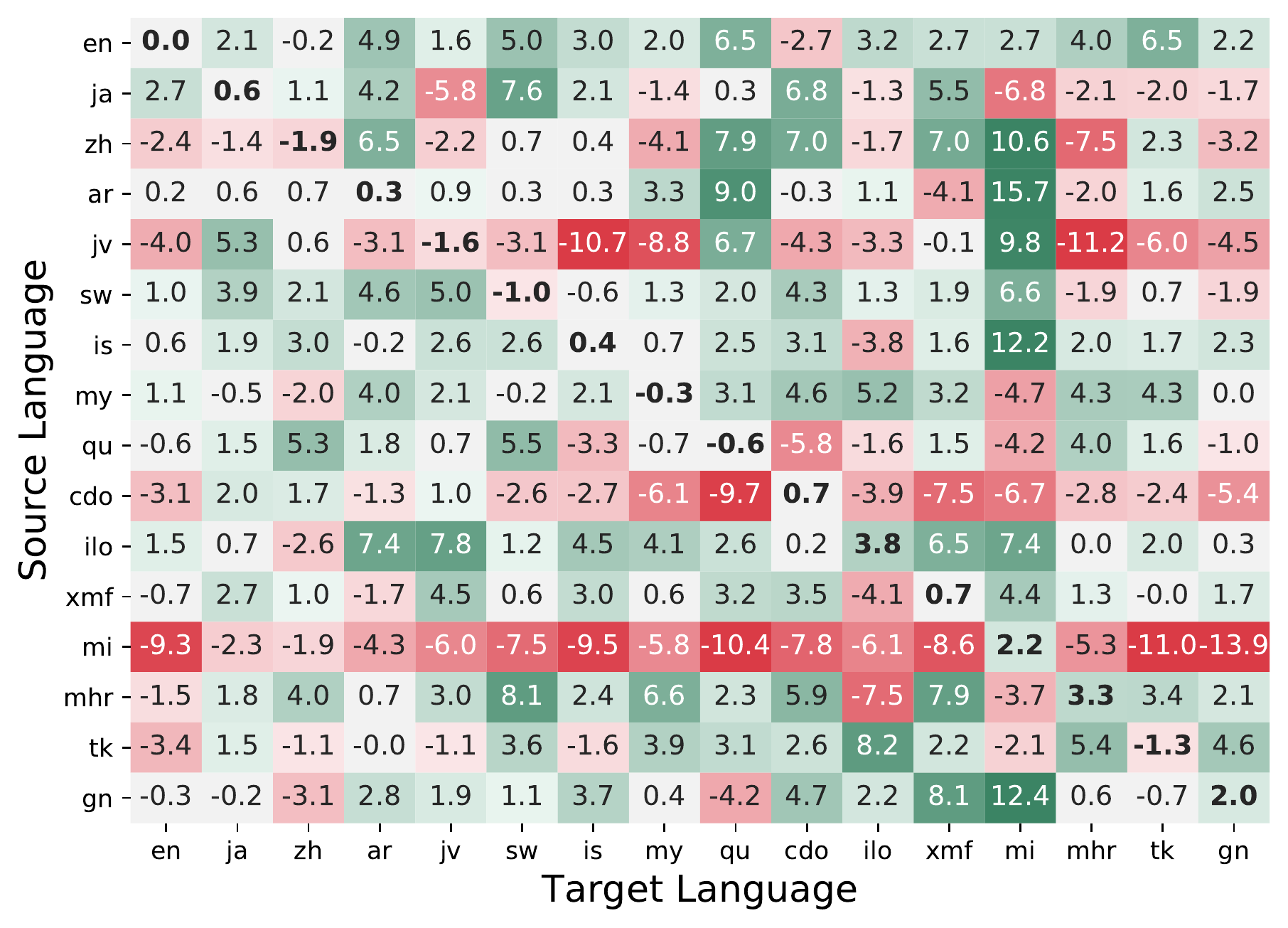}
\caption{Relative $F_1$ improvement of \model{}$^{Base}$ over \model{}$^{Base}$ --\textsc{inv} in cross-lingual NER transfer.}
\label{fig:invvsAd}
\end{figure*}

We show the heatmaps which depict relative F1 improvements of the full \model{}$^{Base}$ framework in the cross-lingual NER transfer task over: (a) the baseline model \mlmtrg{} (Figure~\ref{fig:invvstrg}) and (b) the \model{}$^{Base}$ variant without invertible adapters: \model{}$^{Base}$ --\textsc{inv} (Figure~\ref{fig:invvsAd}).

\begin{figure*}[b] 
\centering
\includegraphics[width=0.9\textwidth]{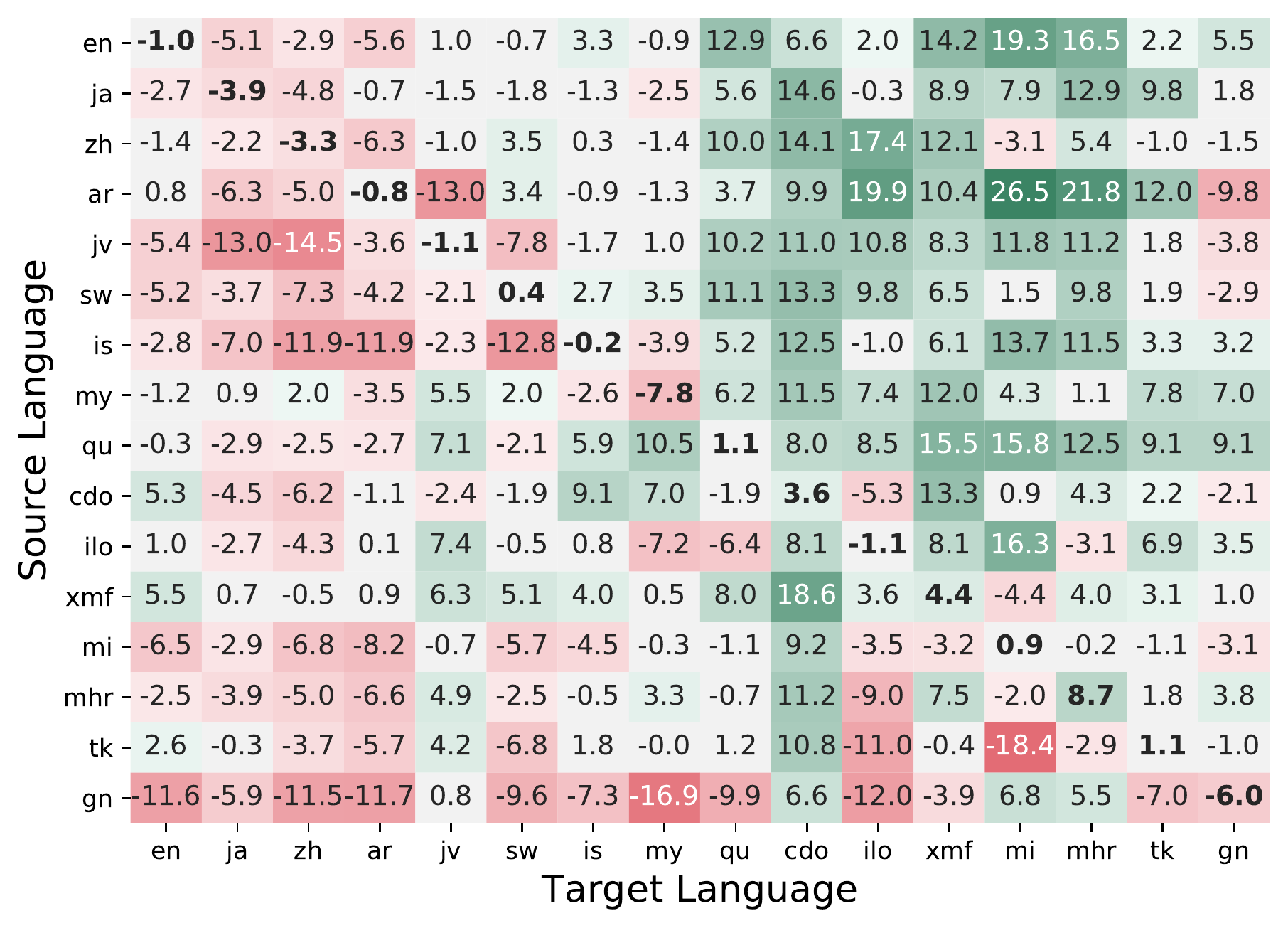}
\caption{Relative $F_1$ improvement of \model{}$^{mBERT}$ over mBERT in cross-lingual NER transfer.}
\label{fig:mbertcomp}
\end{figure*}

The heatmap which depicts relative F1 improvements of the full \model{}$^{mBERT}$ framework   over mBERT can be found in Figure~\ref{fig:mbertcomp}.

\begin{figure*}[b] 
\centering
\includegraphics[width=0.9\textwidth]{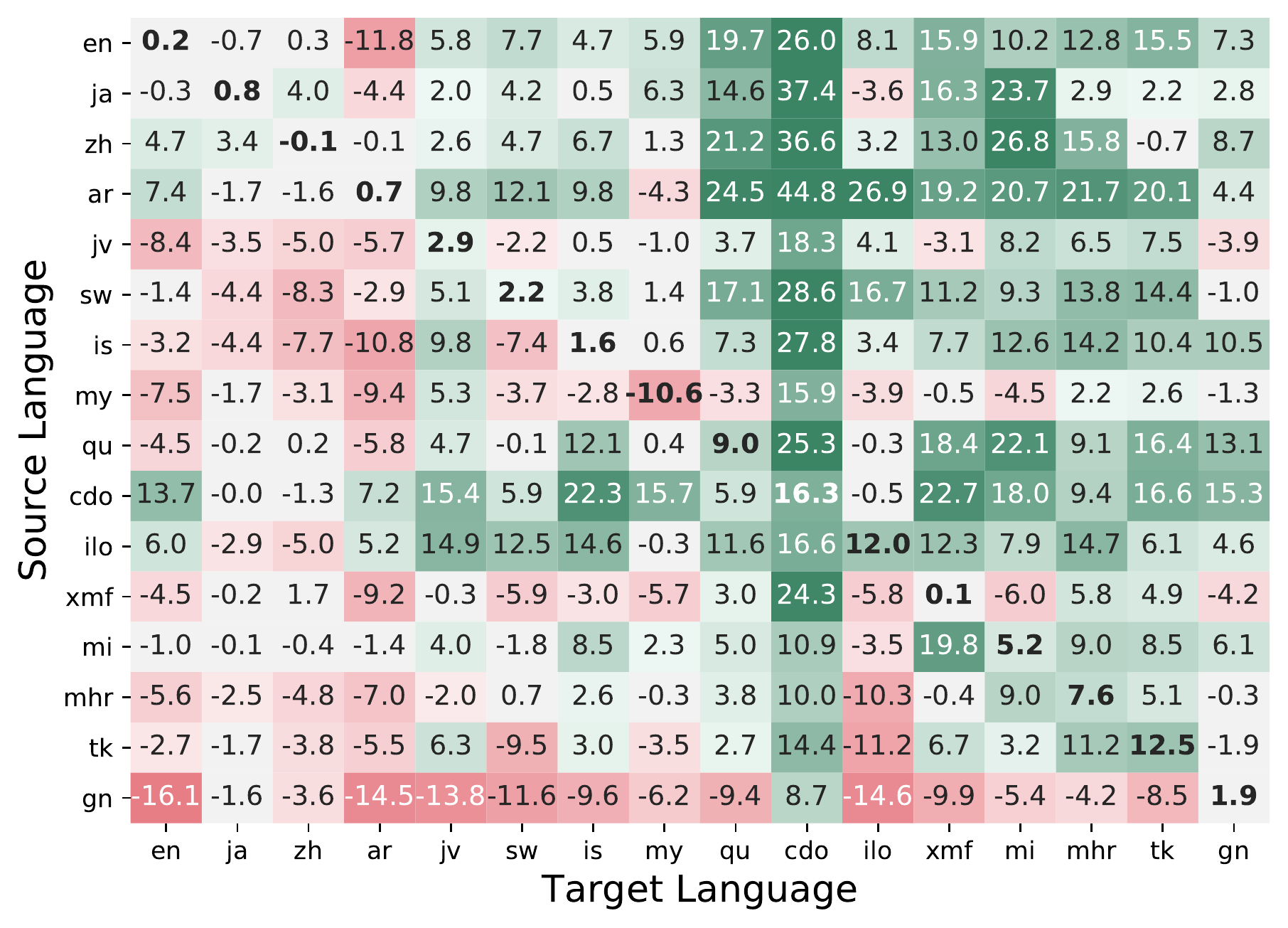}
\caption{Relative $F_1$ improvement of \model{}$^{Large}$ over XLM-R$^{Large}$ in cross-lingual NER transfer.}
\label{fig:xlmrLargecomp}
\end{figure*}

Finally, the heatmap which depicts relative F1 improvements of the full \model{}$^{Large}$ framework   over \xlmr{}$^{Large}$ can be found in Figure~\ref{fig:xlmrLargecomp}.

\section{XCOPA results for all settings}
\label{s:xcopa-all-results}

We show the results on XCOPA for all fine-tuning settings in Table~\ref{tab:xcopa-full-results}.

\begin{table*}[t]
\centering
\resizebox{\textwidth}{!}{%
\begin{tabular}{ll:lllllllllll | l}
\toprule
Model	&	en	&	et	&	ht	&	id	&	it	&	qu	&	sw	&	ta	&	th	&	tr	&	vi	&	zh	&	avg	\\
\midrule
\xlmr{}$^{Base}_{\rightarrow COPA}$	&	57.6	&	59.8	&	49.4	&	58.0	&	56.0	&	50.7	&	57.2	&	56.6	&	52.8	&	56.2	&	58.5	&	56.6	&	55.8	\\
\mlmtrg{}$_{\rightarrow COPA}$	&	57.6	&	57.8	&	48.6	&	60.8	&	54.4	&	49.5	&	55.4	&	55.8	&	54.2	&	54.8	&	57.6	&	57.2	&	55.3	\\
\xlmr{}$^{Base}_{\rightarrow SIQA}$	&	68.0	&	59.4	&	49.2	&	67.2	&	63.6	&	51.0	&	57.6	&	58.8	&	61.6	&	60.4	&	65.8	&	66.0	&	60.7	\\
\xlmr{}$^{Base}_{\rightarrow SIQA \rightarrow COPA}$	&	66.8	&	58.0	&	51.4	&	65.0	&	60.2	&	51.2	&	52.0	&	58.4	&	62.0	&	56.6	&	65.6	&	\textbf{68.8}	&	59.7	\\
\mlmtrg{}$_{\rightarrow SIQA \rightarrow COPA}$ &	66.8	&	59.4	&	50.0	&	\textbf{71.0}	&	61.6	&	46.0	& \textbf{58.8}	&	60.0	&	\textbf{63.2}	&	\textbf{62.2}	&	\textbf{67.6}	&	67.4	&	61.2	\\
\midrule

\model{}$^{Base}_{\rightarrow COPA}$	&	48.1	&	49.0	&	51.5	&	50.7	&	50.7	&	49.1	&	52.7	&	52.5	&	48.7	&	53.3	&	52.1	&	50.4	&	50.7	\\
\model{}$^{Base}_{\rightarrow SIQA}$	&	67.6	&	59.7	&	51.7	&	66.2	&	\textbf{64.4}	&	\textbf{54.0}	&	53.9	&	61.3	&	61.1	&	60.1	&	65.4	&	66.7	&	61.0	\\
\model{}$^{Base}_{\rightarrow SIQA \rightarrow COPA}$	&	\textbf{68.3}	&	\textbf{61.3}	&	\textbf{53.7}	&	65.8	&	63.0	&	52.5	&	56.3	&	\textbf{61.9}	&	61.8	&	60.3	&	66.1	&	67.6	&	\textbf{61.5}	\\
\bottomrule
\end{tabular}%
}
\caption{Accuracy scores of all models on the XCOPA test sets when transferring from English. Models are either only fine-tuned on the COPA training set (${}_{\rightarrow COPA}$), only fine-tuned on the SIQA training set (${}_{\rightarrow SIQA}$) or fine-tuned first on SIQA and then on COPA (${}_{\rightarrow SIQA \rightarrow COPA}$).}
\label{tab:xcopa-full-results}
\end{table*}

\end{document}